%% file: main.tex
\documentclass[10pt,twocolumn,letterpaper]{article}

\usepackage{iccv}              % To produce the CAMERA-READY version
\usepackage{bm}
\usepackage{amsmath,amsthm}
\usepackage{multirow}
\usepackage{array}
\usepackage{tabularray}
\usepackage[percent]{overpic}
\usepackage{array}
\usepackage{subcaption}
\usepackage{arydshln}
\usepackage[ruled,vlined]{algorithm2e}
\usepackage{algorithmic,float}


\definecolor{iccvblue}{rgb}{0.21,0.49,0.74}
\usepackage[pagebackref,breaklinks,colorlinks,allcolors=iccvblue]{hyperref}

\title{Towards Initialization-free Calibrated Bundle Adjustment}

\author{Carl Olsson \quad Amanda Nilsson\\
Lund University\\
}

\def\x{{\bm x}}
\def\U{{\bm U}}

\begin{document}
\maketitle

\begin{abstract}
A recent series of works has shown that initialization-free BA can be achieved using pseudo Object Space Error (pOSE) as a surrogate objective. The initial reconstruction-step optimizes an objective where all terms are projectively invariant and it cannot incorporate knowledge of the camera calibration. As a result, the solution is only determined up to a projective transformation of the scene and the process requires more data for successful reconstruction. 

In contrast, we present a method that is able to use the known camera calibration thereby producing near metric solutions, that is, reconstructions that are accurate up to a similarity transformation.
To achieve this we introduce pairwise relative rotation estimates that carry information about camera calibration. These are only invariant to similarity transformations, thus encouraging solutions that preserve metric features of the real scene. 
Our method can be seen as integrating rotation averaging into the pOSE framework striving towards initialization-free calibrated SfM.

Our experimental evaluation shows that we are able to reliably optimize our objective, achieving convergence to the global minimum with high probability from random starting solutions, resulting in accurate near metric reconstructions. 
\end{abstract}

\section{Introduction}
Bundle adjustment \cite{triggs1999,hartley-zisserman-2003,byrod-eccv-2010,agarwal-etal-eccv-2010,kai-etel-2007,engels-etal-2006} and similar optimization formulations are key components in systems that solve Structure from Motion (SfM) and Simultaneous Localization and Mapping (SLAM) problems \cite{pollefeys-etal-ijcv-2008,Snavely_SIGGRAPH06,agarwal-etal-2011,moulon2013,olsson2011,schoenberger2016sfm,pan2024}. The optimization problem is well known to be non-convex with numerous local minima thus requiring a suitable initialization to converge to the right solution. 

The most common way of achieving this is through incremental reconstruction \cite{pollefeys-etal-ijcv-2008,Snavely_SIGGRAPH06,agarwal-etal-2011,schoenberger2016sfm} where the starting solution is built by adding cameras and 3D points to a two-view-base-reconstruction \cite{nister2004}, by solving consecutive resection and triangulation problems \cite{hartley-sturm-1997,hartley-zisserman-2003}. While errors are typically small in the initial stages of reconstruction the method suffers from error build up, so called drift \cite{cornelis2004}, forcing the incorporation of additional intermediate bundle adjustment stages to reduce reprojection errors.
\begin{figure*}
    \centering
    \begin{overpic}[width=0.33\linewidth]{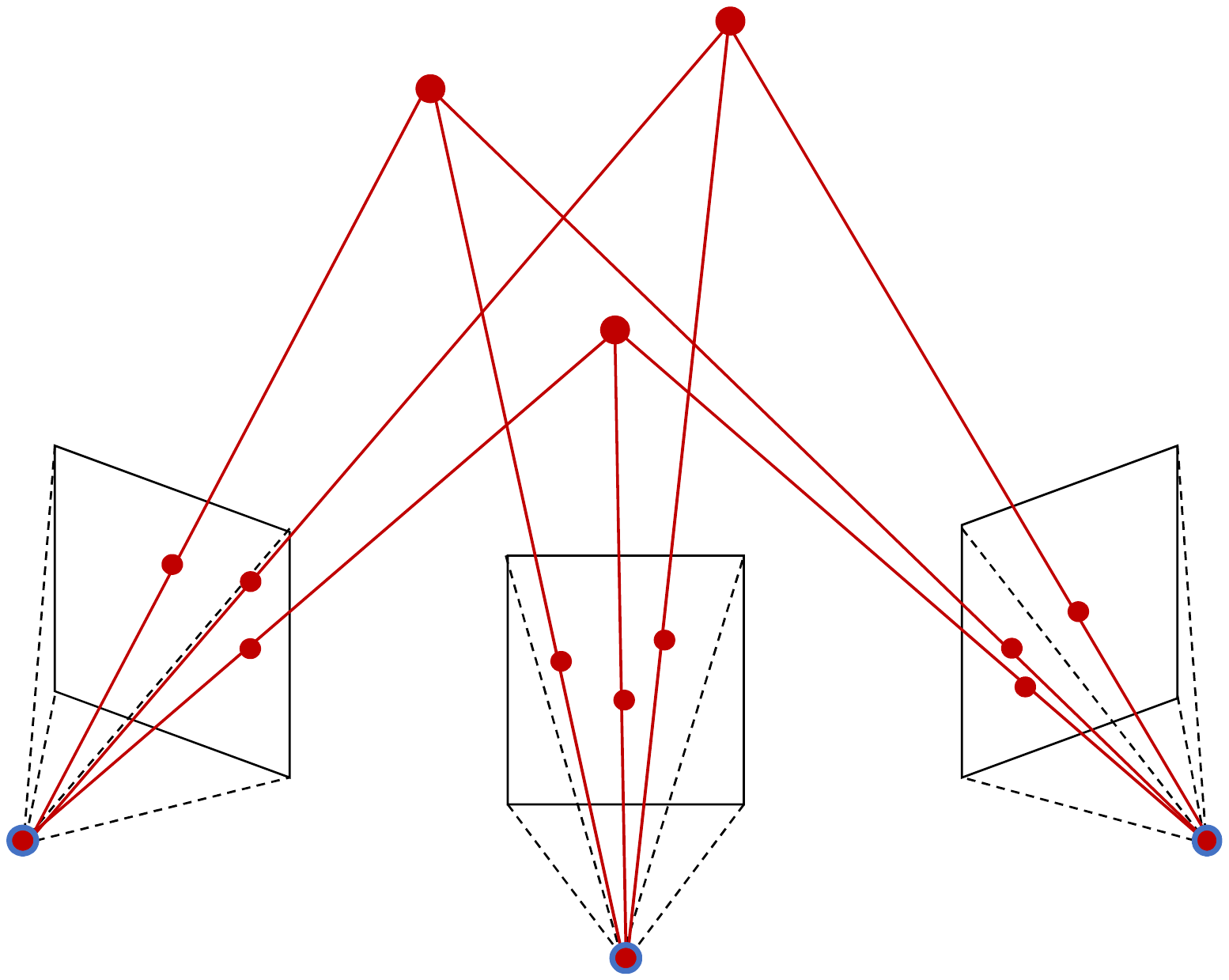}
    \put (27,63){\footnotesize $\bm U_{j}$}
    \put (0,8){\footnotesize $P_{i}$}
    \put (0,45){\footnotesize $P_i \bm U_j \sim \x_{ij}$}
    \end{overpic}
    \begin{overpic}[width=0.33\linewidth]{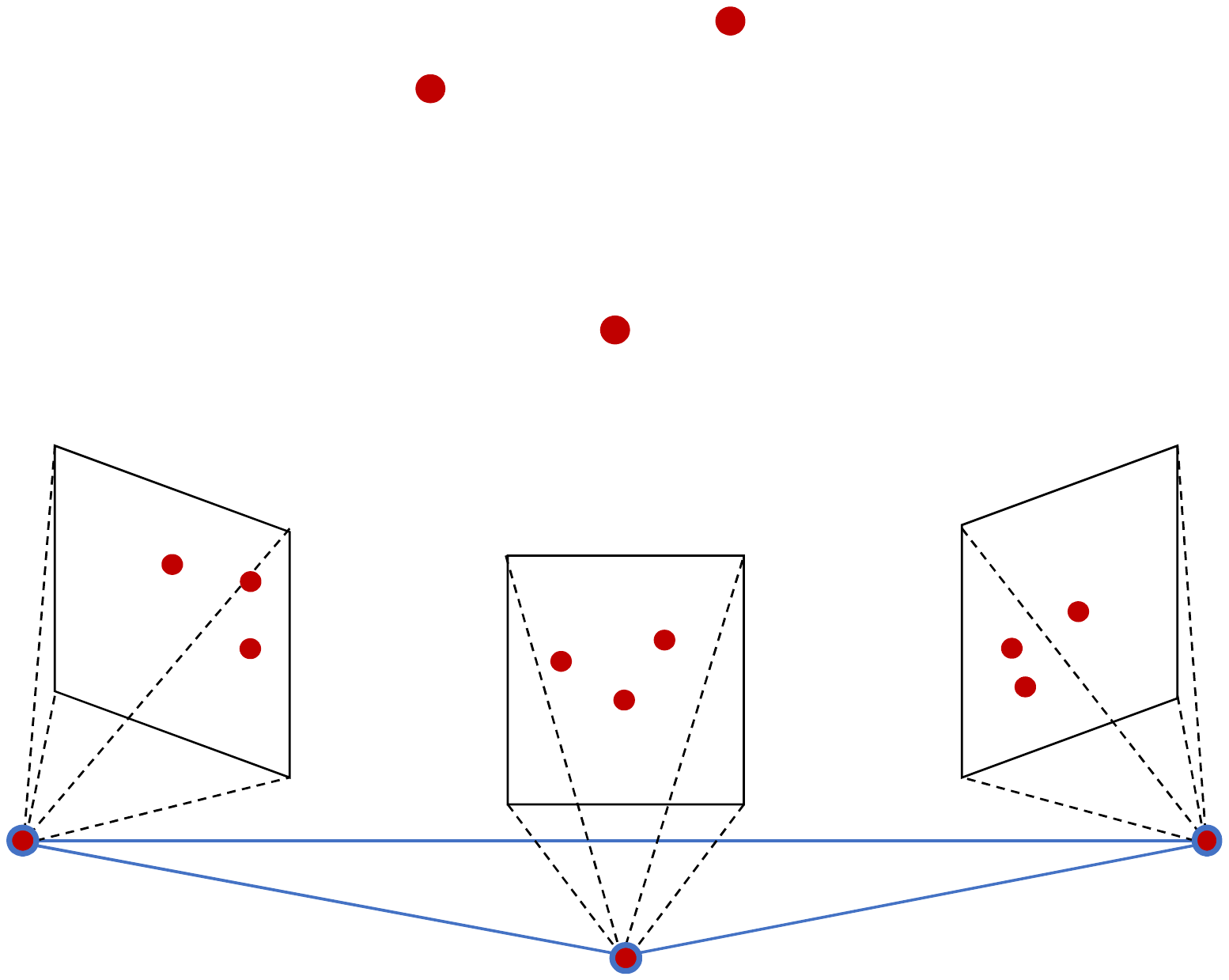}
    \put (27,63){\footnotesize $\bm U_{j}$}
    \put (0,8){\footnotesize $P_i$}
    \put (45,5){\footnotesize $P_k$}
    \put (12,1){\footnotesize $\tilde{R}_{ik}=R_i R_k^T$}
    \end{overpic}
    \begin{overpic}[width=0.33\linewidth]{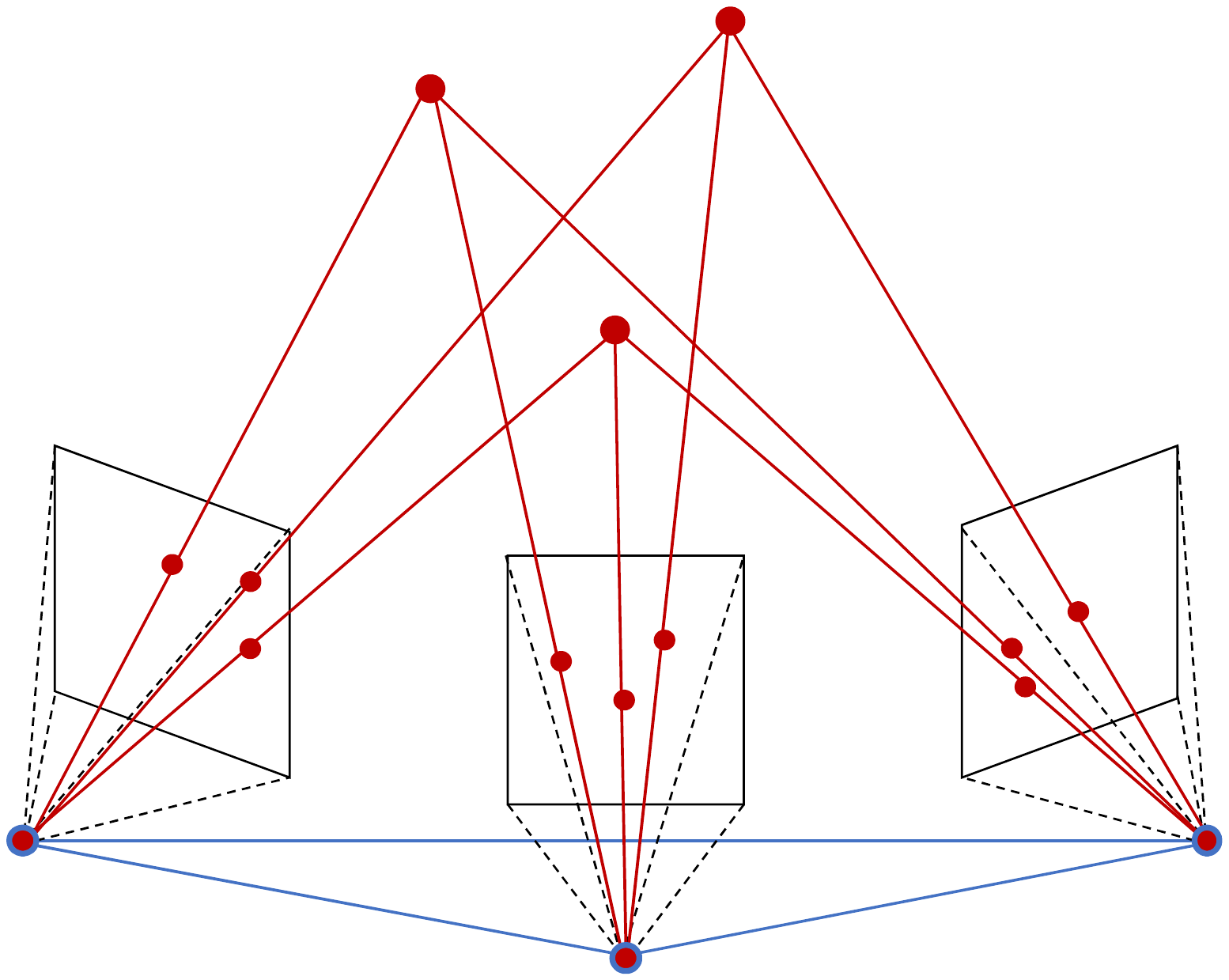}
    \put (27,63){\footnotesize $\bm U_{j}$}
    \put (0,8){\footnotesize $P_{i}$}
    \put (0,45){\footnotesize $P_i \bm U_j \sim \x_{ij}$}
    \put (45,5){\footnotesize $P_k$}
    \put (12,1){\footnotesize $\tilde{R}_{ik}=R_i R_k^T$}
    \end{overpic}
    \caption{\emph{Left:} Bipartite bundle adjustment graph. Edges (red) correspond to reprojection errors of observed point projections, which are invariant to projective 3D transformation. \emph{Middle:} Rotation averaging graph. Edges (blue) correspond to relative rotation errors which are invariant to similarity transformations. \emph{Right:} Our method uses both reprojection and relative rotation errors to achieve a pOSE formulation which is only invariant to similarity transforms.}
    \label{fig:camera_graphs}
\end{figure*}

Non-sequential methods (sometimes referred to as global methods) \cite{martinec-pajdla-2007,olsson2011,enqvist2011,arie-nachimson-etal-2012,moulon2013,pan2024} attempt to use as much data as possible in each stage of the reconstruction. The key component is the so called \textit{rotation averaging} or \textit{synchronization problem} \cite{dai2009,hartley2011,singer2011,fredriksson2012,hartley2013,wang2013,wilson2016,briales2017,Carlone2015b,chatterjee2017,dellaert2020,wilson2020,chitturi2021,Moreira2021,parra2021,chen2021,eriksson2021,Zhang2023}. This can be thought of as optimization over the camera graph, where nodes correspond to unknown camera orientations and edges between nodes represent relative rotation estimates between neighboring cameras. In contrast, regular bundle adjustment can be seen as optimization over a bipartite graph, where nodes correspond to both cameras and 3D points, see Figure~\ref{fig:camera_graphs}.
The relative rotations are typically obtained by solving the two-view-relative-pose \cite{nister2004} for each possible pair of images with more than five image correspondences. 
The formulation is able to handle the non-convex rotation constraints of calibrated cameras through convex relaxation to a linear semi-definite programming problem \cite{eriksson2021}. 
It has been shown both empirically \cite{wilson2016} and theoretically \cite{eriksson2021,rosen2021} that if the camera graph is sufficiently dense the problem has no other local minima than the global one. 
After the absolute camera rotations have been determined, camera translations can be estimated through translation averaging \cite{kennedy-etal-2012,moulon2013,cui-etal-bmvc-2015,cui-tan-iccv-2015,hainan-etal-2016,latit-govindu-2022} and 3D point positions can be triangulated. 
Alternatively, camera locations and 3D point positions can be determined simultaneously by solving the so called \textit{known rotation problem} \cite{rother-carlsson-2001,rother-carlsson-eccv-2002,kahl-hartley-tpami-2008,olsson-etal-cvpr-2010,zhang-etal-2018} (which has recently been re-branded as \textit{Global Positioning} \cite{pan2024}). Since rotation averaging only optimizes over camera parameters it gives a significant reduction of the number of parameters compared to the full bundle adjustment problem. On the other hand, the input to the method is computed from independently solved two-view relative pose problems. This means that 3D points and camera positions may not be consistently estimated in all two-view problems. Therefore, refinement with bundle adjustment may still be a requirement to achieve low reprojection errors.

A recent line of works have explored the possibility of achieving initialization-free bundle adjustment through the use of pseudo object space errors (pOSE) \cite{hong_2018,hong-etal-eccv-2016,iglesias-olsson-iccv-2021,Iglesias_2023_CVPR,weber-etal-eccv-2024}. It has been demonstrated that these methods achieve convergence to the global minimum from random starting solutions with high probability \cite{hong_2018}. They handle calibration through a stratified approach \cite{pollefeys1997stratified,faugeras19983,faugeras1995stratification,hartley-1999,fusiello-ivc-2000,hartley-zisserman-2003}. First an initial uncalibrated reconstruction is computed. The resulting reconstruction is then upgraded, by applying a projective transformation to cameras and 3D points, to achieve camera-matrices with given internal parameters. This results in a so called metric reconstruction which is related to the true 3D geometry through a similarity transformation \cite{hartley-zisserman-2003}. We note however that there are certain problems with this approach:
Firstly, the upgrade problem is overdetermined and there is no guarantee that there is a projective transformation that gives the right internal parameters for all cameras, unless the setting is noise-free. In practice it is often observed that linear upgrade methods can give complex parameters in the presence of noise, yielding un-realizable solutions.
Secondly, since uncalibrated cameras have more degrees of freedom (DOF) than calibrated ones (11 vs. 6) the initial uncalibrated reconstruction requires more data to give stable results.

In this paper we propose an optimization formulation that finds near metric reconstructions. 
By making camera matrices conform to estimated relative rotation matrices between pairs of cameras we reduce the projective invariance to a similarity transformation. 
Our approach fuses the pOSE and rotation averaging frameworks, see Figure~\ref{fig:camera_graphs}, giving a formulation that is insensitive to initialization while incorporating knowledge of intrinsic camera parameters in the objective. 
We do not explicitly enforce the given camera intrinsics, since this makes the optimization sensitive to local minima, but with our extra penalty terms solutions that have roughly correct calibration is encouraged effectively fusing the bundle adjustment and upgrade steps into one. 
In summary, our main contributions are:
\begin{itemize}
    \item We show how the projective invariance of bundle adjustment methods can be removed by incorporating relative rotation estimates. 
    \item We present a new objective function that incorporates both pOSE errors, ensuring good reconstruction quality, as well as pairwise camera penalties, that encourage solutions to be near metric.
    \item We show that the resulting optimization problem can be solved using the VarPro algorithm \cite{hong_2017} and verify experimentally that this results in reliable inference, converging to the global optimum from random starting solutions with high probability.
\end{itemize}

\section{Projective Ambiguity and Upgrades}
In this Section we give a short review of projective ambiguity in uncalibrated SfM and the traditional way of resolving this through upgrades. We then present our approach for resolving this ambiguity through the inclusion of relative rotation estimates. 

\subsection{Uncalibrated Reconstruction}\label{sec:uncalib}

Given a number of 2D image projections $\x_{ij}$, 
the goal of uncalibrated reconstruction is to find camera matrices $P_i$ and 3D points $\U_i$ so that 
\begin{equation}
    \lambda_{ij} \x_{ij} = P_i \U_j,
\end{equation}
for some non-zero numbers $\lambda_{ij}$.
Here $\x_{ij}$ is a 3D vector representing the 2D projection, $P_i$ is a $3 \times 4$ matrix representing the camera and $\U_{j}$ is a 4D vector representing the 3D projection, see \cite{hartley-zisserman-2003} for details on these representations. We remark that these quantities are homogeneous, meaning that the matrix/vector represents the same camera/point after scaling.
The scale-factor $\lambda_{ij}$ is often referred to as the projective depth of point $j$ in camera $i$.

The equations have multiple solutions. By introducing a projective 3D transformation represented by a $4\times 4$ matrix $H$ a new solution $\tilde{P}_i = P_i H$, $\tilde{\U}_j = H^{-1}\U_j$ with the same projections is introduced, since $P_i \U_j = \tilde{P}_i \tilde{\U}_i$.
Algebraically, we can also change the scale of $P_i$ and $\U_j$, and compensate by changing the projective depth $\lambda_{ij}$, although from a projective point of view this is still the same solution.
The projective reconstruction theorem \cite{nasihatkon-etal-2015} states that (under some mild technical conditions) all solutions are projectively equivalent, meaning that they are all related by a projective transformation as described above.

We can think of the uncalibrated reconstruction problem in terms of a camera-point graph, see Figure~\ref{fig:camera_graphs}. 
The nodes of the graph consists of cameras and 3D points that are to be estimated, while the edges connecting the nodes represents a reprojection error. This graph is bipartite and all the edge costs are invariant to a projective transformation.

\subsection{Upgrades}
General projective reconstruction often appear visually unreasonable since the application of an unknown projective transformations does not preserve lengths or angles and may even place visible points behind the camera or at infinity. To achieve visually plausible reconstructions the solution can be upgraded to conform to prior knowledge of the internal camera parameters. 
If $P_i = K_i \begin{bmatrix}
    R_i & t_i
\end{bmatrix}$, where the $3 \times 3$ matrix $K_i$ contains the inner parameters \cite{hartley-zisserman-2003}, $R_i$ is a rotation matrix representing the camera orientation and $t_i$ is a vector controlling its position, we can pre-multiply the image points $\x_{ij}$ with $K_i^{-1}$, which restricts the camera matrices to be of the form $\tilde{P}_i = \begin{bmatrix}
    R_i & t_i
\end{bmatrix}$ (which we will refer to as calibrated cameras). Thus, from an uncalibrated reconstruction the upgrade step seeks to find an invertible $4\times 4$ matrix $H$ such that $P_i H_{1:3} \sim R_i$, where $\sim$ means equality up to scale and $H_{1:3}$ are the first 3 columns of $H$. With $\Omega = H_{1:3} H_{1:3}^T$ we get
\begin{equation}
P_i \Omega P_i^T \sim R_i R_i^T = I.
\label{eq:upgrade}
\end{equation}
which leads to an overdetermined linear homogeneous system. Upgrade methods typically solve the above equations for $\Omega$, in a least squares sense, and then extract $H$.

\subsection{Using Relative Rotation Measurements}
While \eqref{eq:upgrade} leads to a simple algorithm it is however unlikely that an exact solution exists in the presence of noise. In addition, the extraction of $H$ from $\Omega$ may have complex results that have to be modified to obtain a realizable solution. Moreover, an inexact upgrade step does not take reprojection error into account when modifying the cameras, and therefore typically degrades the quality of the solution.

In this work, we want to add constraints to the initial reconstruction step to ensure that the solution we get is close to calibrated without upgrading it. 
In principle, one could directly add the quadratic $R_k R_k^T = I$ in the optimization (which also selects the scale of the camera), or alternatively use a direct rotation parametrization (e.g. with skew-symmetric matrices and the matrix exponential). However, as we shall see in the experimental evaluation (see Section~\ref{sec:locminexp}) this leads to methods that frequently get stuck in very poor local minima. 

The reason that terms of the form $R_k R_k^T$ can be used for upgrades is that they are not invariant under the projective transformation. We therefore seek to include other non-invariant penalties in the objective function.
We observe that in general the measured relative rotations $\tilde{R}_{kl}$ between cameras are not invariant since 
\begin{equation}
P_k \Omega P_l^T \sim R_k R_l^T  = \tilde{R}_{kl}
\end{equation}
will constrain $\Omega$. 
We will therefore add penalties of the form 
\begin{equation}
\ell^{rot}_{kl}(R_k R_l^T) = \left\|\sqrt{W_{kl}} \operatorname{vec}\left(R_kR_l^T-\tilde{R}_{kl}\right)\right\|^2.
\label{eq:relroterror}
\end{equation}
Here, $R_k$ is the first $3 \times 3$ part of the camera $P_k$. The matrix $\sqrt{W_{kl}}$ is the square root of a precision matrix that has two functions. Firstly, it is designed to penalize deviations from the tangent plane of the rotation manifold at the estimate $R_{kl}$. Secondly, it makes \eqref{eq:relroterror} approximate changes in reprojection errors in the two-view problem where $\tilde{R}_{kl}$ was originally estimated, hence encouraging values of $R_k R_l^T$ that do not increase errors in the two view problem much compared to $R_{kl}$.
(For more details on the construction of $\sqrt{W}$ see Section~\ref{sec:W}.)

The new terms can be seen as introducing additional edges between camera nodes of the camera-point graph, see Figure~\ref{fig:camera_graphs}. It has been observed in the context of pure rotation averaging that if the camera nodes are densely connected relative rotation costs result in well behaved problems where local minima rarely occur. We will see that when replacing regular reprojection costs with pOSE costs we obtain a well behaved problem that can be solved reliably.

\subsection{Invariance of the Fundamental Matrix}\label{Section2.4}
We conclude this section by deriving a measure of how close to upgradable a solution is.
%Since the process of upgrading a solution may involve both non-linear steps and approximations we next present an easy-to-test condition that ensures that a solution to the uncalibrated reconstruction problem can/cannot be upgraded. 
Given camera matrices $P_k = \begin{bmatrix}
    A_k & t_k
\end{bmatrix}$, $k=1,...,f$ we let the fundamental matrices be $F_{kl} \sim A_l^{-T}[c_k - c_l]_\times A_k^{-1}$, where $c_k = -A_k^{-1} t_k$ are camera centers. These matrices all have two non-zero singular values, and in the special case that $A_k$ are rotations the two non-singular values are the same.
The fundamental matrix $F_{ij}$ is uniquely defined up to scale by the requirement that $P_k^T F_{kl} P_l$ 
should be skew symmetric \cite{hartley-zisserman-2003}, meaning that 
\begin{equation}
    \U^T P_k^T F_{kl} P_l \U = 0,
    \label{eq:skewsym}
\end{equation}
for all 4D vectors $\U$. 
Since $P_k\U = P_kHH^{-1}\U = \tilde{P}_k \tilde{\U}$ it is clear by \eqref{eq:skewsym} that the fundamental matrix is invariant to the projective transformation $H$. Hence, if there is a fundamental matrix $F_{kl}$ that has two different non-zero singular values then there is no projective transformation $H$ that changes all $A_i$ to rotations. 
The converse, that there is such an $H$ if all $F_{kl}$ are essential matrices, is also true for cameras whose centers are not collinear \cite{kasten_2019_ICCV}. (In such cases the set of essential matrices determine the camera matrices uniquely up to a projective transformation, and it can be seen that there is a solution with only calibrated camera matrices.)
Since fundamental matrices are scale invariant we use the normalized difference $\frac{\sigma_1-\sigma_2}{\sigma_1+\sigma_2}$, where $\sigma_1 \geq \sigma_1 > 0$ are the singular values of a fundamental matrix, as a measure of how close the matrix is to being essential.

\section{pOSE with Relative Rotation Estimates}
In this Section we present our approach for combining the pOSE framework with rotation averaging. We first give a brief overview of the pOSE framework and then show how to add relative rotation estimates so that the resulting formulation can be effectively solved using 2nd order methods such as VarPro \cite{hong-fitzgibbon-iccv-2015,hong_2017}.

\subsection{pOSE}
\begin{figure}
    \centering
    \includegraphics[width=0.9\linewidth,trim={4.5cm 9cm 4.5cm 9cm},clip]{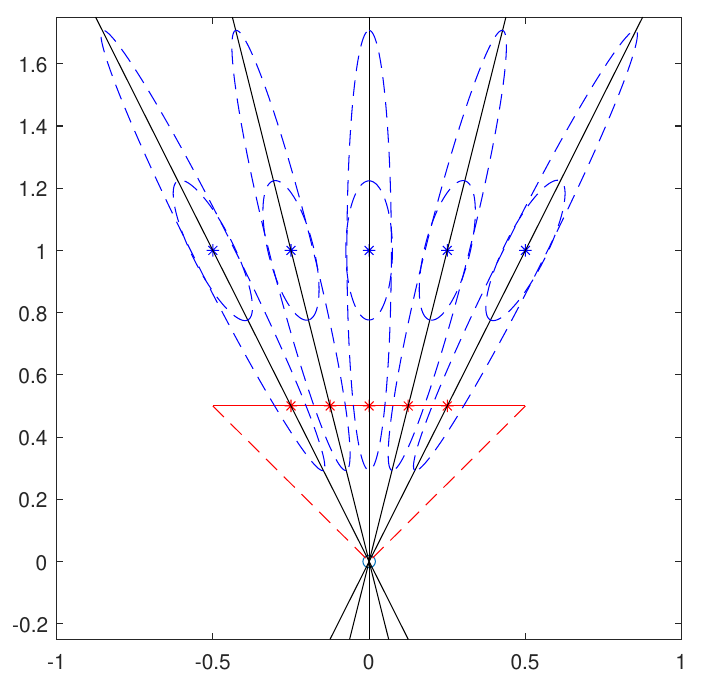}
    \caption{Levelsets for (a 2d version of) the pOSE objective for $\eta = 0.1$ and $0.01$ for a five different projections.}
    \label{fig:levelsets}
\end{figure}

The pOSE framework \cite{hong_2018} was introduced as an alternative to bundle adjustment which is much less dependent on the quality of the initial solution. It has been empirically demonstrated to converge to the right solution with high probability from random starting solutions. The main difference to regular bundle adjustment is that the nonlinear perspective error is replaced with the object space error. If $m_{ij}$ is a 2D vector, representing the observed projection of the 3D point $\U_j$ in camera $P_i$ then the reprojection error is
\begin{equation}
    \left\|m_{ij} - \frac{P_i^{1:2}\U_j}{P_i^{3}\U_j}\right\|^2,
    \label{eq:reproj}
\end{equation}
where $P_i^{1:2}$ contains the first two rows of $P_i$ and $P_i^3$ the third.
The pOSE frameworks switches this term for 
\begin{equation}
  \ell^{OSE}_{ij}(P_i\U_j) =  \left\|m_{ij}P_i^{3}\U_j - P_i^{1:2}\U_j\right\|^2.
\end{equation}
The above term can be seen to measure a distance between the 3D line that projects to $m_{ij}$ and the point $P_i \U_j$. In contrast to \eqref{eq:reproj} it is however homogeneous meaning that the trivially setting $P_i = 0$ and $\U_j = 0$, which is not projectively meaningful, gives a zero penalty. 
For this reason, one adds weak terms that penalize solutions with projective depths being zero. In this work we use 
\begin{equation}
\ell^{aff}_{ij}(P_i \U_j) = \left(\frac{m_{ij}^TP_i^{1:2}\U_j+P_i^{3}\U_j}{\|m_{ij}\|^2 + 1} - 1 \right)^2.
\end{equation}
This term measures the distance between $P_i \U_j$ and $(m_{ij},1)$ along the 3D line projecting to $m_{ij}$.
It will favor solutions that have projective depth that is approximately $1$, hence effectively preventing trivial solutions. The pOSE framework minimizes the convex combination $\ell_{pOSE}(\{P_i\},\{\U_i\}) =$
\begin{equation}
     \sum_{ij} (1-\eta) \ell^{OSE}_{ij}(P_i\U_j)  + \eta \ell^{aff}_{ij}(P_i \U_j),
\label{eq:pOSE}
\end{equation}
where $\eta$ is a parameter that is typically selected small to allow large variations in projective depth.
Figure~\ref{fig:levelsets} shows examples of level sets of the pOSE errors for five different projections.
Here $\eta = 0.1$ and $0.01$. Smaller values of $\eta$ allow larger variations of the projective depth.
The main benefit of using the pOSE framework is that the objective function is a bilinear least squares problem in the unknowns $\{P_i\},$ and $\{\U_j\}$. Thus, unlike regular reprojection error, there is a closed form solution for the 3D points $\{\U_j\}$ in terms of the camera parameters $\{P_i\}$. This allows the use of VarPro \cite{hong_2017} which has been shown to greatly improve global convergence properties compared to methods such as Gauss-Newton or Levenberg-Marquardt \cite{hong-fitzgibbon-iccv-2015}.

\subsection{pOSE with Rotation Averaging}\label{sec:W}
Next we describe how we incorporate estimates of relative rotations in the pOSE framework.
The relative rotation estimate $\tilde{R}_{kl}$ between two cameras $k$ and $l$ are obtained by solving two-view relative pose. For ease of notation we will drop the indices $k,l$ in this section. Since the two-view problem is invariant to a global similarity transformation \cite{hartley-zisserman-2003} we can assume that
the first camera is 
$\begin{bmatrix}
    I & 0
\end{bmatrix}$ and the second is $\begin{bmatrix}
    R & t
\end{bmatrix}$, with $R \in SO(3)$ and $\|t\|^2=1$.
The objective thus depends on $R$, $t$ and the parameters of the 3D points. We let $v$ be a vector containing $t$ and the parameters of all 3D points visible in the two cameras.
Further, we let $r(R,v)$ be a vector containing all reprojection residuals and write $\|r(R,v)\|^2$ for the sum-of-squared reprojection errors.
Close to a local minimum $(\tilde{R},\tilde{v})$ the function can be approximated by
\begin{equation}
\| \tilde{r} + J\xi+ K\Delta v\|^2 = \|\tilde{r}\|^2 + \|J\xi+K \Delta v\|^2.
\label{eq:jac}
\end{equation}
Here $\tilde{r} = r(\tilde{R},\tilde{v})$ and $J$ and $K$ are Jacobians with respect to rotation parameters $\xi$ and the remaining parameters $v$ respectively. The equality above comes from the fact that at a local minimum the gradients, $J^T \tilde{r}$ and $K^T \tilde{r}$ with respect to $\xi$ and $\Delta v$ respectively, vanish.
The vector $\xi$ is 3D vector containing the angle-axis representation of a rotation that changes the locally optimal $\tilde{R}$ according to $e^{[\xi]_\times}\tilde{R}$.

Since \eqref{eq:jac} is quadratic, minimizing with respect to the parameters in $\Delta v$ gives $\Delta v = -K^\dagger J \xi$, which inserted into \eqref{eq:jac} gives
\begin{equation}
    \|(I-KK^{\dagger})J \xi\|^2 = \xi^T J^T (I-KK^{\dagger}) J \xi.
    \label{eq:axisangle}
\end{equation}
Since the above minimization determines the best camera position and 3D points for any choice of $\xi$ the above expression can be seen to locally approximate the best sum-of-squares reprojection error that we can achieve for a given choice of rotation. In our formulation, we wish to avoid the non-linear terms involved when using direct rotation parametrization. Since we do not strictly enforce rotation constraints we need to "lift" \eqref{eq:axisangle} so that it can be applied to a general $3 \times 3$ matrix (or to its vectorization).
For this purpose we choose $3$ orthonormal $3\times 3$ matrices $B_i = \frac{1}{\sqrt{2}}[e_i]_\times$, $i=1,2,3$, where $e_i$ is column $i$ of the $3\times 3$ identity matrix. We complement these to an orthonormal basis by selecting the $6$ matrices $B_i$, $i=4,..,9$ to be symmetric and orthonormal.
The matrices $V_i := \frac{1}{\sqrt{3}}B_i \tilde{R}$, $i=1...9$ then also constitute an orthonormal basis of $\mathbb{R}^{3\times 3}$ where the first three matrices span the tangent-space of the rotation manifold at $\tilde{R}$ and the last $6$ the normal space. 
We now introduce the orthogonal $9 \times 9$ matrix 
\begin{equation}
V = \begin{bmatrix}
    \operatorname{vec}(V_1) & \hdots & \operatorname{vec}(V_9)
\end{bmatrix}.
\end{equation}
To 'lift' \eqref{eq:axisangle} we observe that a general matrix $\Delta R$, can be written $ \Delta R = (S+[\xi]_\times)\tilde{R}$, for some symmetric $S$ and angle-axis vector $\xi$. 
By construction, the first three elements of the vector $V^T\operatorname{vec}(\Delta R)$ will contain $\sqrt{2}\xi$ while the last six contain the coefficients for writing $S$ as a linear combination of the $B_4,...,B_9$. 
Thus if we let $A$ be a $9 \times 9$ matrix who's top left $3 \times 3$ block is $\frac{1}{2}J^T(I-KK^\dagger)J$ we obtain a matrix $W = V A V^T$ that fulfills
\begin{equation}
     \operatorname{vec}([\xi]_\times\tilde{R})^T W \operatorname{vec}([\xi]_\times\tilde{R}) = \xi^T J^T (I-KK^{\dagger}) J \xi,
\end{equation}
which means that the "lifted" penalty agrees with \eqref{eq:axisangle} for any matrix that is in the tangent plane.
To encourage solutions on the rotation manifolds we also introduce a penalty for deviations in normal directions.

This can be done in many different ways. A simple way which we use is to let the lower right $6 \times 6$ block of $A$ be an identity matrix. This penalizes deviations in the normal directions uniformly. 
Since $J^T (I-KK^{\dagger})J$ is (symmetric and) positive semi-definite it is easy to see that so will $W = VAV^T$ be.
Thus we can find its unique positive semi-definite square root $\sqrt{W}$ (through for example eigendecomposition).
Letting $\Delta R = R-\tilde{R}$ we arrive at a penalty of the form 
\begin{equation}
    \operatorname{vec}(R-\tilde{R})^T W \operatorname{vec}(R-\tilde{R}) = \|\sqrt{W}\operatorname{vec}(R-\tilde{R})\|^2.
\end{equation}
For each camera pair $k,l$ (where two view relative pose problem is solvable) we obtain an estimate $\tilde{R}_{kl}$ and a matrix $\sqrt{W_{kl}}$. Thus if $P_k = \begin{bmatrix}
    R_k & t_k
\end{bmatrix}$ and $P_l = \begin{bmatrix}
    R_l & t_l
\end{bmatrix}$ we penalize deviations of the relative camera rotation $R_k R_l^T$ from $\tilde{R}_{kl}$ using \eqref{eq:relroterror}. If $k=l$ we we use $\tilde{R}_{kl} = \sqrt{W_{kl}} = I$.
This gives the objective function
\begin{equation}
     \ell_{pOSE}({P_i},{U_j}) + \beta \sum_{k,l} \ell^{rot}_{kl}(R_k R_l^T).
     \label{eq:rotpOSE}
\end{equation}
where $\beta$ controls the tradeoff between pOSE and rotation averaging. In this paper we use $\beta = 1$ for all datasets. We remark that we do not introduce any weighting for the last term. We find that since both terms approximately measure reprojection error this works well.

\section{Optimization Method}
In this section we describe the method that we use to optimize \eqref{eq:relroterror}. 
To simplify notation we define the block matrices
\begin{equation}
B = \begin{bmatrix}
    R_1 \\ R_2 \\ \vdots
\end{bmatrix}, \  
t = \begin{bmatrix}
    t_1 \\ t_2 \\ \vdots
\end{bmatrix} \text{ and }
C^T = \begin{bmatrix}
    u_1 & u_2 & \hdots 
\end{bmatrix}.
\end{equation}
Here $u_i$ are regular Cartesian 3D coordinates for point $i$.
If $\bm 1$ is a vector of ones then $X = BC^T+t\bm 1^T$ is the block matrix that contain all the vectors $P_i \U_j$. Similarly $Y = BB^T$ is the block matrix that contain all matrices $R_i R_i^T$.
All terms of the objective function \eqref{eq:rotpOSE} are linear least squares terms in these matrices allowing us to write it as 
\begin{equation}
    \|\mathcal{L}_{rot}(Y)-b_{rot}\|^2+\|\mathcal{L}_{pOSE}(X)-b_{pOSE}\|^2.
    \label{eq:matpoSErot}
\end{equation}
Since all the objectives are linear least squares problems in $C$ and $t$ it is possible to find optimal values for these in closed form for any given $B$. We denote these optimal value functions $C^*(B)$ and $t^*(B)$. Conceptually, the VarPro algorithm \cite{hong_2017} inserts back into \eqref{eq:matpoSErot} giving an objective that only depends on $B$.
The resulting optimization problem is then locally optimized.
It can be shown \cite{hong_2017} that this is the same as running Levenberg-Marquardt on the full problem, without any dampening on the $C$ and $t$ terms, with an extra (exact update) of $C$ and $t$ in each step.
To describe the process we let $b = \operatorname{vec}(B)$, $v = \begin{bmatrix} \operatorname{vec}(C) \\t \end{bmatrix}$ and 
$r(b,v)$ be the vector of all residual errors so that the objective \eqref{eq:matpoSErot} can then be written $\|r(b,v)\|^2$. Locally around a point $(\tilde{b},\tilde{v})$ the objective can be approximated by 
\begin{equation}
    \|r(b,v)\|^2 \approx \|J \Delta b + K \Delta v + \tilde{r}\|^2,
    \label{eq:gnapprox}
\end{equation}
where $J$ and $K$ are Jacobians of $r(b,v)$ with respect to $b$ and $v$ respectively, and $\tilde{r} = r(\tilde{b},\tilde{v})$.
Minimizing with respect to $\Delta v$ gives $\Delta v = - K^\dagger (J \Delta b + \tilde{r})$ which inserted into \eqref{eq:gnapprox} 
\begin{equation}
    \|P(J\Delta b + \tilde{r})\|^2,
    \label{eq:gnapprox2}
\end{equation}
where $P$ is the projection matrix $(I-KK^\dagger)$.
VarPro solves this problem repeatedly with a dampening term $\lambda \|\Delta b\|^2$, and updates $\tilde{b} = \tilde{b} + \Delta b$. After each update of $\tilde{b}$ an additional (exact) update of $\tilde{v}$ is performed. We summarize the method in Algorithm~\ref{alg:varpro}\footnote{The code will be made publicly available from the first authors homepage.}.

\begin{algorithm}[t]
\SetAlgoLined
\KwResult{Optimal b,v}
 Choose initial $\tilde{b}$ and $\tilde{v}$ randomly\;
 Compute \textit{error} $=\|r(\tilde{b},\tilde{v})\|^2$\;
 \While{not converged}{
 Compute $\tilde{r}$, $J$ and $K$ at $(\tilde{b},\tilde{v})$\;
Set $\bar{b} \leftarrow \tilde{b} + \operatorname{argmin} \|P(J\Delta b + \tilde{r})\|^2+\lambda\|\Delta b\|^2$\;
Recompute $\tilde{r}, K$ around $(\bar{b},\tilde{v})$\;
  Set $\bar{v} \leftarrow \tilde{v}-K^\dagger\tilde{r}$\;
  
  \eIf{error $>\|r(\bar{b},\bar{v})\|^2$ }{
  Update $\tilde{b} \leftarrow \bar{b}$, 
   $\tilde{v} \leftarrow \bar{v}$, and 
   
   \textit{error} $\leftarrow \|r(\bar{b},\bar{v})\|^2$\;
   $\lambda \leftarrow \lambda/1.25$\;
   }{
   $\lambda \leftarrow 10 \lambda$\;
  }
 }
 \caption{VarPro for solving \eqref{eq:matpoSErot}.}
 \label{alg:varpro}
\end{algorithm}

\section{Experiments}
To evaluate the effects of incorporating rotation averaging in the pOSE framework we have test the pOSE model using the objective $l_{pOSE}$ \eqref{eq:pOSE} and our proposed method incorporating rotation averaging by adding the $l_{rot}$ penalty to $l_{pOSE}$, giving the objective \eqref{eq:rotpOSE}. %The $l_{rot}$ term is a pairwise camera constraint that encourages rotation matrix products $R_iR_j^T$ to be close to their corresponding relative rotations $R_{ij}$. 

For comparison we also test two modifications. The first one only adds a penalty $l_{diag} = \sum_i \|R_i R_i^T-I\|^2$, that measures the deviations from the orthogonal constraints, to the $l_{pOSE}$. The second one uses direct rotation parametrization, to strictly enforce rotation constraints, as described in \ref{sec:W} with the objective \eqref{eq:rotpOSE}. It is only presented in combination with the pairwise camera term $l_{rot}$ is because it gets stuck in local minima too frequently without it. 

\subsection{Effects of different rotation constraints}\label{sec:locminexp}
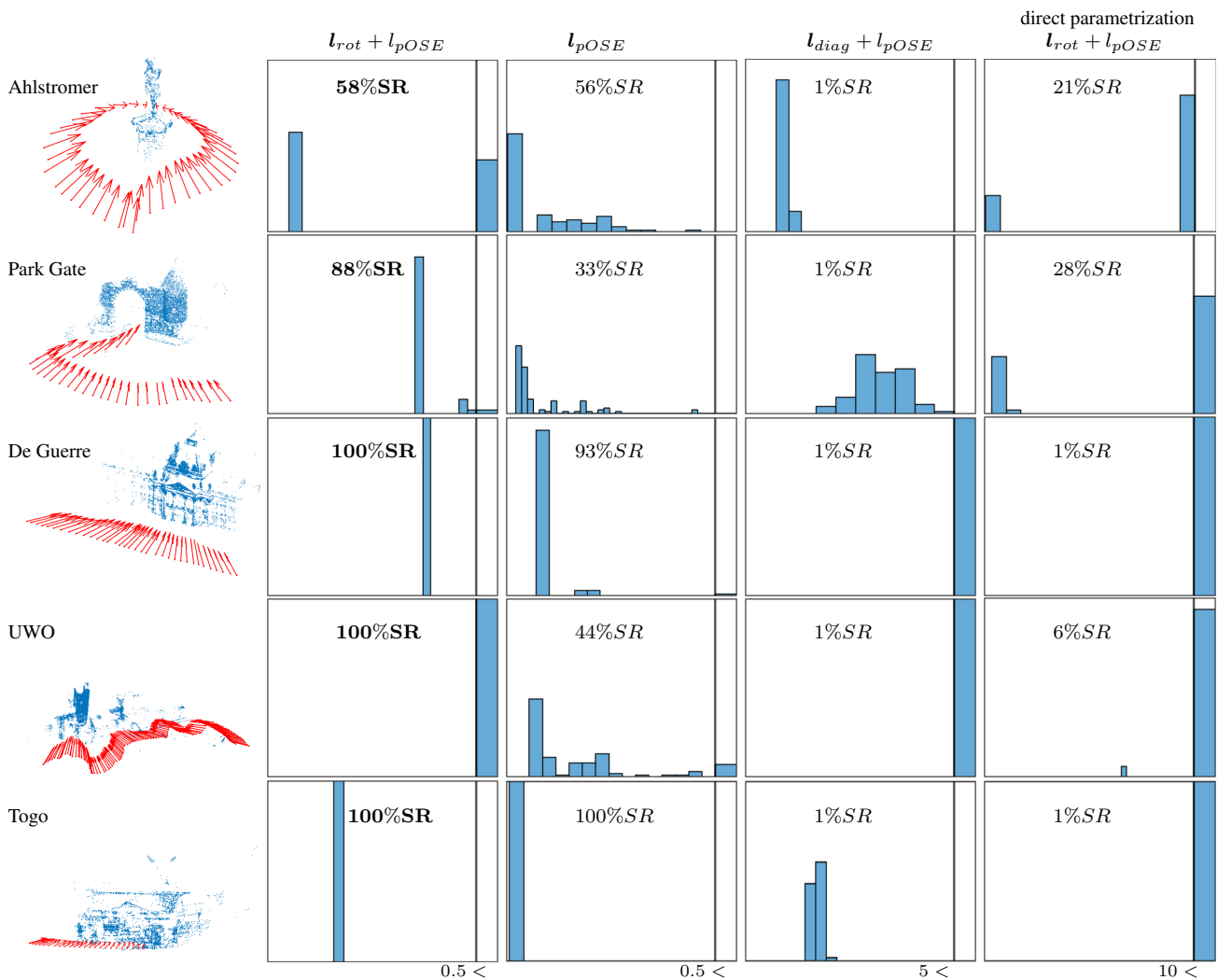
\begin{figure*}[h]
    \centering
    \begin{overpic}[width=0.19\linewidth]{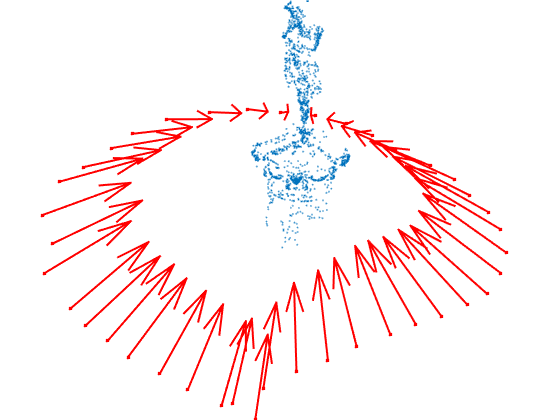}
    \put (-8,60){\footnotesize Ahlstromer}
    \end{overpic}
    \begin{overpic}[width=0.19\linewidth]{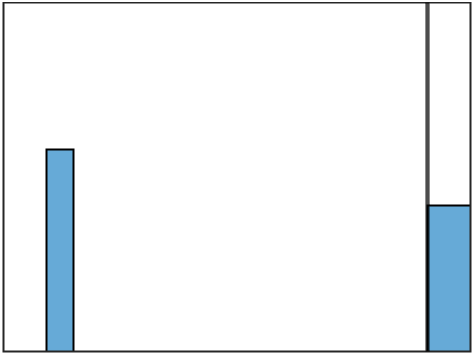}
    \put (27,80){\footnotesize $\bm l_{rot}+l_{pOSE}$}
    \put (30,60){\footnotesize $\bf{58\% SR}$}
    \end{overpic}
    \begin{overpic}[width=0.19\linewidth]{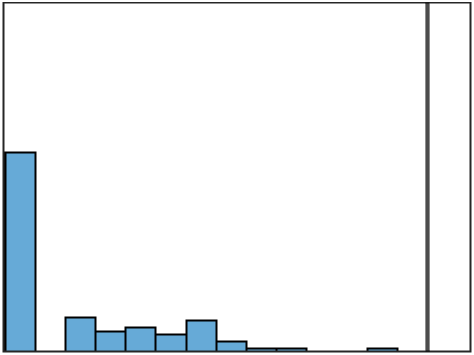}
    \put (27,80){\footnotesize $\bm l_{pOSE}$}
    \put (30,60){\footnotesize $56\% SR$}
    \end{overpic}
    \begin{overpic}[width=0.19\linewidth]{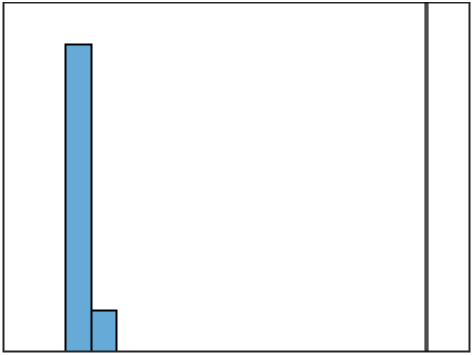}
    \put (27,80){\footnotesize $\bm l_{diag}+l_{pOSE}$}
    \put (30,60){\footnotesize $1\% SR$}
    \end{overpic}
    \begin{overpic}[width=0.19\linewidth]{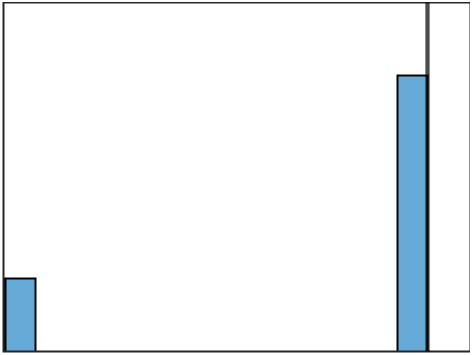}
    \put (27,80){\footnotesize $\bm l_{rot}+l_{pOSE}$}
    \put (16,90){\footnotesize direct parametrization}
    \put (30,60){\footnotesize $21\% SR$}
    \end{overpic}
    
    \begin{overpic}[width=0.19\linewidth]{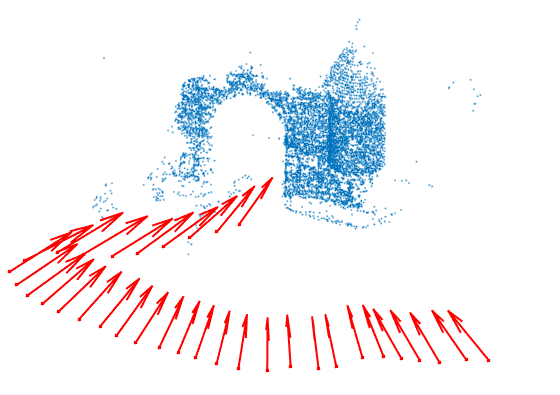}
    \put (-8,60){\footnotesize Park Gate}
    \end{overpic}
    \begin{overpic}[width=0.19\linewidth]{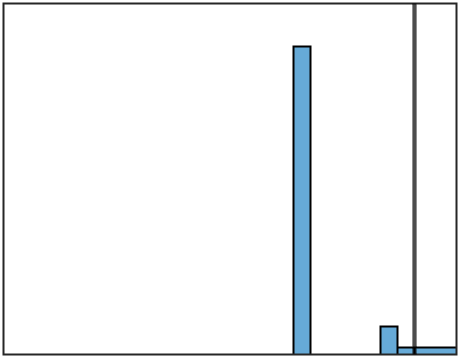}
    \put (28,60){\footnotesize $\bf{88\% SR}$}
    \end{overpic}
    \begin{overpic}[width=0.19\linewidth]{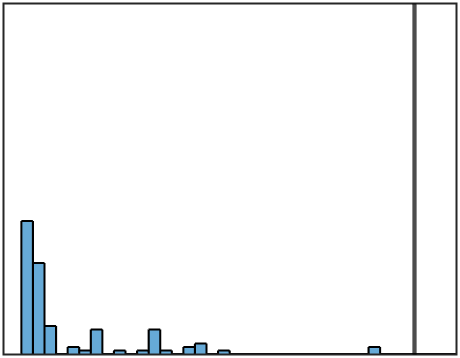}
    \put (30,60){\footnotesize $33\% SR$}
    \end{overpic}
    \begin{overpic}[width=0.19\linewidth]{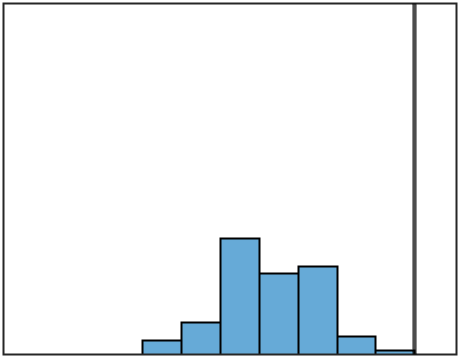}
    \put (30,60){\footnotesize $1\% SR$}
    \end{overpic}
    \begin{overpic}[width=0.19\linewidth]{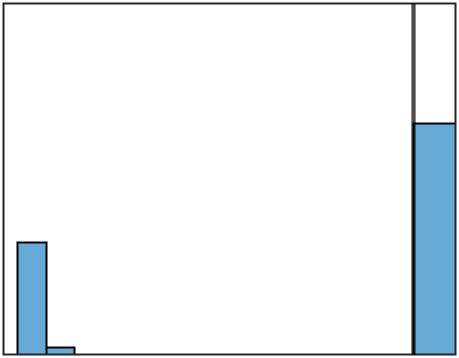}
    \put (30,60){\footnotesize $28\% SR$}
    \end{overpic}

    \begin{overpic}[width=0.19\linewidth]{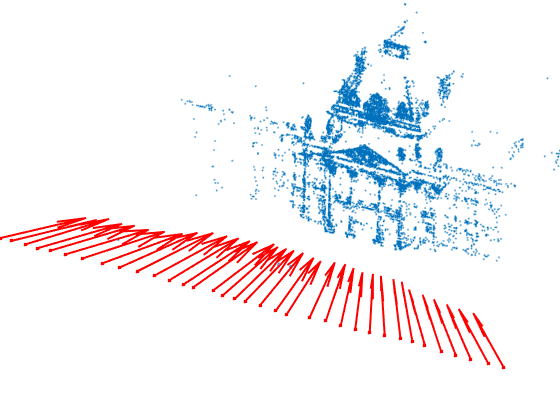}
    \put (-8,60){\footnotesize De Guerre}
    \end{overpic}
    \begin{overpic}[width=0.19\linewidth]{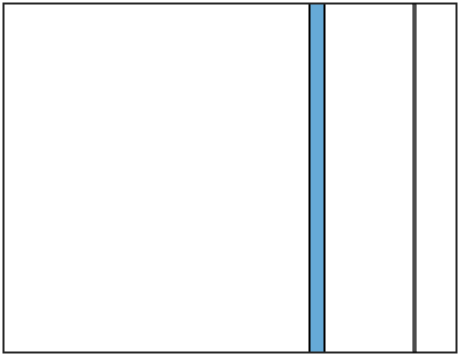}
    \put (28,60){\footnotesize $\bf{100\% SR}$}
    \end{overpic}
    \begin{overpic}[width=0.19\linewidth]{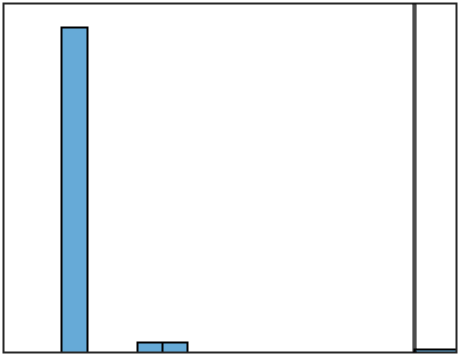}
    \put (30,60){\footnotesize $93\% SR$}
    \end{overpic}
    \begin{overpic}[width=0.19\linewidth]{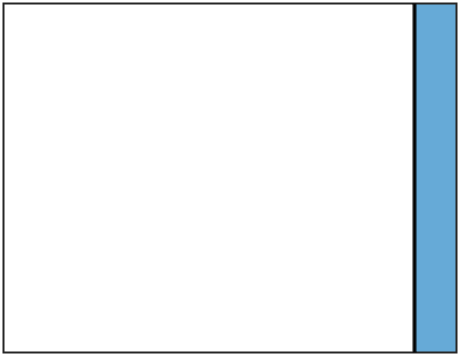}
    \put (30,60){\footnotesize $1\% SR$}
    \end{overpic}
    \begin{overpic}[width=0.19\linewidth]{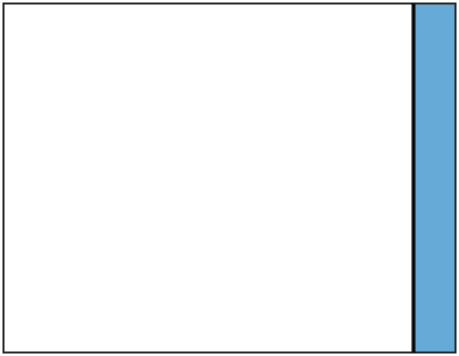}
    \put (30,60){\footnotesize $1\% SR$}
    \end{overpic}

    \begin{overpic}[width=0.19\linewidth]{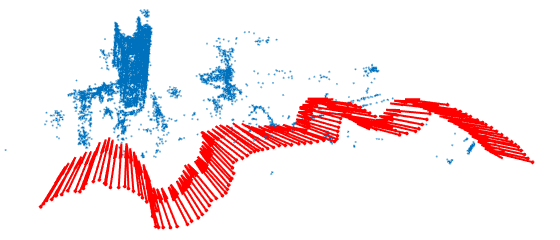}
    \put (-8,60){\footnotesize UWO}
    \end{overpic}
    \begin{overpic}[width=0.19\linewidth]{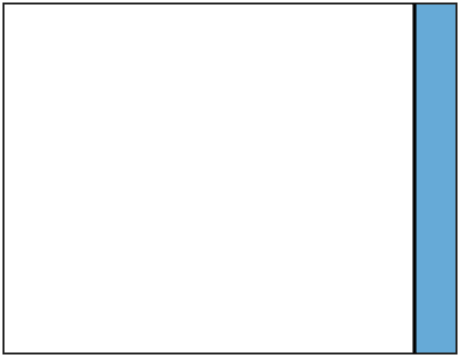}
    \put (30,60){\footnotesize $\bf{100\% SR}$}
    \end{overpic}
    \begin{overpic}[width=0.19\linewidth]{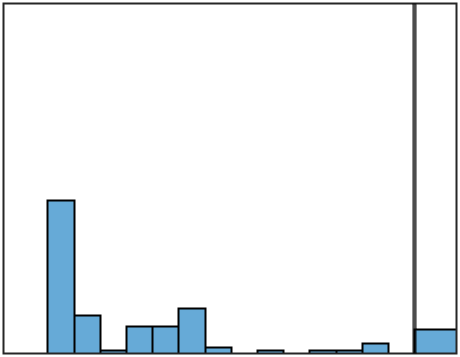}
    \put (30,60){\footnotesize $44\% SR$}
    \end{overpic}
    \begin{overpic}[width=0.19\linewidth]{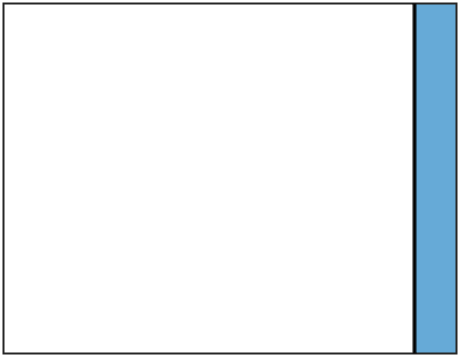}
    \put (30,60){\footnotesize $1\% SR$}
    \end{overpic}
    \begin{overpic}[width=0.19\linewidth]{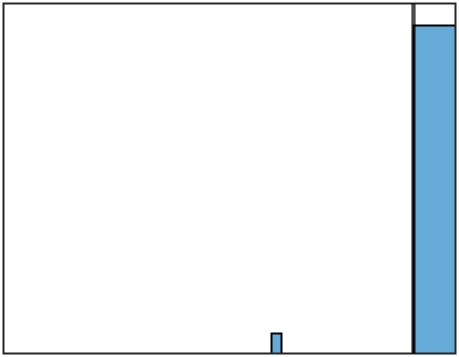}
    \put (30,60){\footnotesize $6\% SR$}
    \end{overpic}

    \begin{overpic}[width=0.19\linewidth]{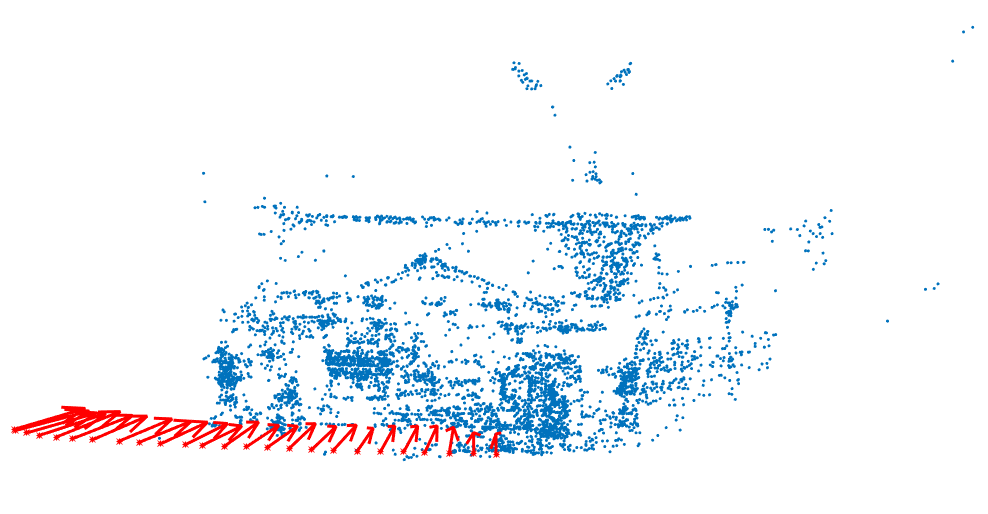}
    \put (-8,60){\footnotesize Togo}
    \end{overpic}
    \begin{overpic}[width=0.19\linewidth,height=2.6cm]{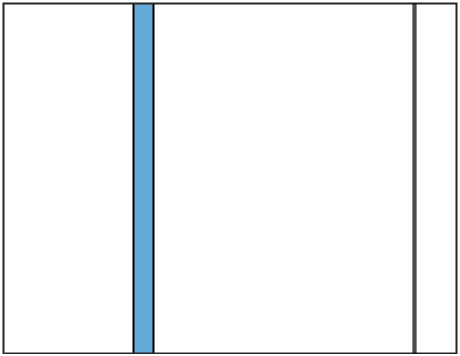}
    \put (35,60){\footnotesize $\bf{100\% SR}$}
    \put (75,-8){\raisebox{2pt}{\scriptsize $0.5<$}}
    %\put (65,-8){\scriptsize $0.5<l$}
    %\put (75,-8){\raisebox{2pt}{\scriptsize $0.5<l$}}
    \end{overpic}
    \begin{overpic}[width=0.19\linewidth,height=2.6cm]{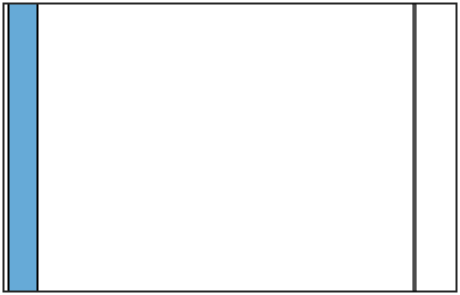}
    \put (30,60){\footnotesize $100\% SR$}
    \put (75,-8){\raisebox{2pt}{\scriptsize $0.5<$}}
    %\put (75,-8){\scriptsize $0.5<l$}
    \end{overpic}
    \begin{overpic}[width=0.19\linewidth,height=2.6cm]{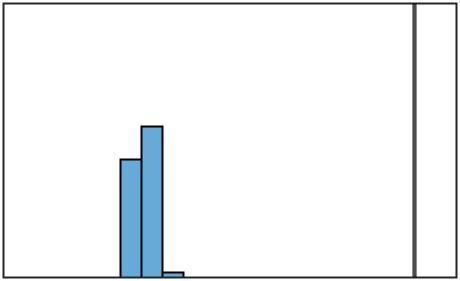}
    \put (30,60){\footnotesize $1\% SR$}
    \put (75,-8){\raisebox{2pt}{\scriptsize $5<$}}
    %\put (75,-8){\scriptsize $5<l$}
    \end{overpic}
    \begin{overpic}[width=0.19\linewidth,height=2.6cm]{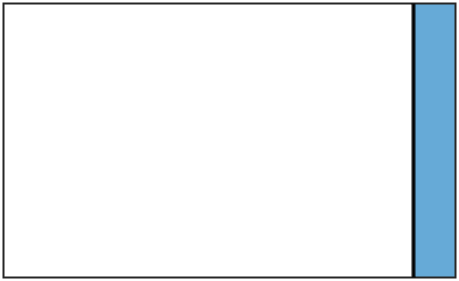}
    \put (30,60){\footnotesize $1\% SR$}
    \put (75,-8){\raisebox{2pt}{\scriptsize $10<$}}
    \end{overpic}
    %\vspace{5pt} 
    \caption{Histograms of local minima for the different methods and their corresponding objective functions to illustrate how often the global optimum is reached. Each row corresponds to the specified dataset whose 3D points and cameras are visualized to the left. All objective values larger than the specified value at the x-axis, marked with a vertical line, is stored in the bin to the right of that line.}
    \label{fig:histogram_fval}
\end{figure*}

In this section we evaluate the proposed methods robustness to local minima. 
For each of the four methods 100 different starting solutions were randomly drawn from a standard normal distribution and the resulting convergence points were evaluated after at most 200 iterations. Histograms of the objective values from these runs on different datasets of varying size can be found in Figure \ref{fig:histogram_fval}. 

Our rotation averaging approach with pairwise camera penalties converges to the global minima with higher probability than the other methods. In particular we notice that the naive approaches of directly parameterizing with rotations and only penalizing deviations from orthogonality has a low success rate (SR)\footnote{Success rate is here the defined as the fraction of 100 runs that reached the minimum of the objective up to a tolerance of $10^{-5}$.}. These methods often converge to local minima very far from the optimum. During these experimental runs it was also noted that our rotation averaging optimization required far less iterations to reach its global minimum - around a third of the iterations needed for the pOSE model to converge \footnote{To reach the optimum with the Ahlstromer dataset, our model only needed  on average 35 iterations (5 s), while the pOSE model required 137 iterations (17s). For Park Gate, 36 iterations (13 s) for our model vs. pOSE’s 83 iterations (30 s); De guerre: 34 iterations (28 s) vs. 100 iterations (81 s); UWO: 35 iterations (72 s) vs. 155 iterations (306 s); and Togo: 36 iterations (5 s) vs. 93 iterations (11 s). One iteration of our and the pOSE formulation take roughly the same time.}. 

We conclude that optimization with the orthogonal constraint and direct parametrization has a very low success rate and tends to get stuck in local minima very far from the global optimum.   

\subsection{Near Metric Reconstructions}
To evaluate how close to essential the fundamental matrices $F_{kl}$, corresponding to the cameras $P_k$ and $P_l$, we use the normalized difference of its singular values $\big(\frac{\sigma_1 - \sigma_2}{\sigma_1 + \sigma_2}\big)_{kl}$. The mean and the range (from smallest to largest), denoted $m(\frac{\sigma_1 - \sigma_2}{\sigma_1 + \sigma_2}\big)$ and $r(\frac{\sigma_1 - \sigma_2}{\sigma_1 + \sigma_2}\big)$ respectively, over all these fundamental matrix measures are used to determine how close to metric the solutions are.
Here we only consider regular pOSE \eqref{eq:pOSE} and pOSE with relative rotation penalties \eqref{eq:rotpOSE}. We remark that while the other two presented approaches provide metric or near metric solutions when successful, their success relies on a good starting solution and we therefore don't consider them further. 

The quality of the projections is evaluated with the pseudo object space error $l_{pOSE}$ since both methods include this error in their respective objectives. The results can be found in Table \ref{Table_result}.
%The results from all experiments can be found in Table~\ref{Table_result}. 
%{\color{red} Next we evaluate how close to metric the solution ... }
%{\color{red}  ....}
%We further evaluated the pOSE method with and without our rotation averaging implementation on larger datasets. 
%The pOSE error $l_{pOSE}$ is used for comparison since the models have different objectives.
As expected the formulation \eqref{eq:rotpOSE} which includes relative rotation penalties $l_{rot}$ will generally produce higher pOSE errors $l_{pOSE}$ than \eqref{eq:pOSE} which only minimizes the pOSE component $l_{pOSE}$. 
%However, for the proposed method the pOSE errors do not increase nearly as much with data size compared to \eqref{eq:pOSE}. The errors are between 5 and 20 times larger for the smaller datasets, but only 2-3 times larger for the largest datasets, with UWO being an exception. 

Our proposed formulation \eqref{eq:rotpOSE} yields solutions that are closer to metric than those returned when only minimizing \eqref{eq:pOSE}. In general the average and range of the fundamental matrix metrics increase with the size of the dataset, especially so when only optimizing the pOSE objective \eqref{eq:pOSE}. See e.g. the dataset Urban where the normalized difference of the singular values range is almost 1. Thus while some fundamental matrices might be close to essential, at least one is very far from it.

When the rotation averaging is applied, all solutions are very close to upgradable, even for the largest datasets. For the smallest datasets with (25 to 35 cameras) the fundamental matrices measures are quite low also for \eqref{eq:pOSE} - only two to four times larger than those attained with our rotation averaging approach, but for larger datasets they become between 6 and over 50 times larger. Even for the midsize dataset Ahlstromer with 41 cameras the mean and range become over 20 and 15 times the size of the ones for the estimates from our rotation averaging optimization.

\setlength\extrarowheight{3pt}
\begin{table*}[h]
	\def\w{35mm}
	\begin{center}
		\begin{tabular}{cc|ccccc} %H to hide the column
            \hline
            \hline
		\multicolumn{2}{c}{Dataset:}	  &Objective & $l_{pOSE}$ &$\big(\frac{\sigma_1 - \sigma_2}{\sigma_1 + \sigma_2}\big)$ & $r\big(\frac{\sigma_1 - \sigma_2}{\sigma_1 + \sigma_2}\big)$\\
			\hline

\multirow{3}{*}{\begin{minipage}{\w}
    Sri Mariamman \\
    \textit{Cameras: \hspace{0.15cm} 222  \\
    3D points: \hspace{0.01cm} 40 658}
\end{minipage}} & \multirow{3}{*}{\includegraphics[height=14mm]{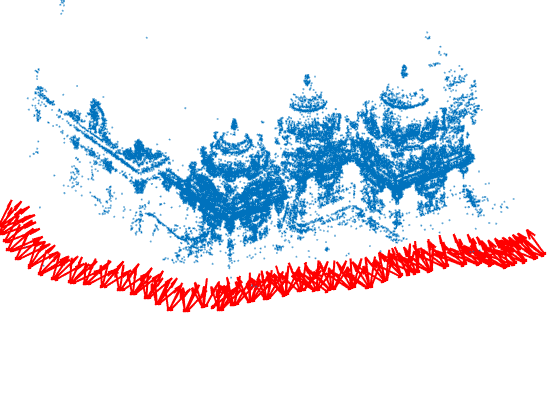}} & \eqref{eq:pOSE} & \textbf{1.289 } & 0.036 & 0.537 &\\
&& \eqref{eq:rotpOSE} &  3.57 &  \textbf{0.005} & \textbf{0.033}\\ 
& & \\
\hline

\multirow{3}{*}{\begin{minipage}{\w}
    Urban II \\
    \textit{Cameras: \hspace{0.15cm} 90  \\
    3D points: \hspace{0.01cm} 22 205}
\end{minipage}} & \multirow{3}{*}{\includegraphics[height=14mm]{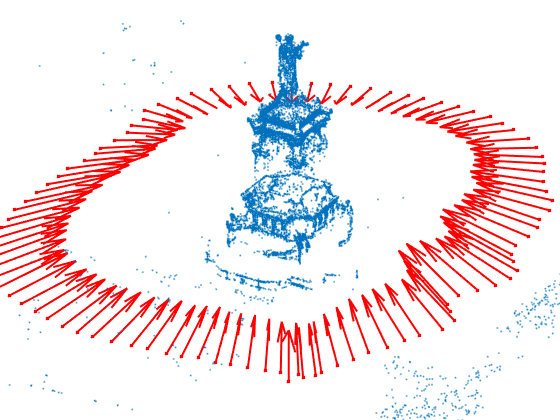}} & \eqref{eq:pOSE} & \textbf{ 1.916} & 0.208 & 0.978 &\\
&& \eqref{eq:rotpOSE} & 4.898 & \textbf{0.004} & \textbf{0.022 }\\ 
& &\\
\hline

\multirow{3}{*}{\begin{minipage}{\w}
    Court Yard \\
    \textit{Cameras: \hspace{0.15cm} 241  \\
    3D points: \hspace{0.01cm} 48 643}
\end{minipage}} & \multirow{3}{*}{\includegraphics[height=14mm]{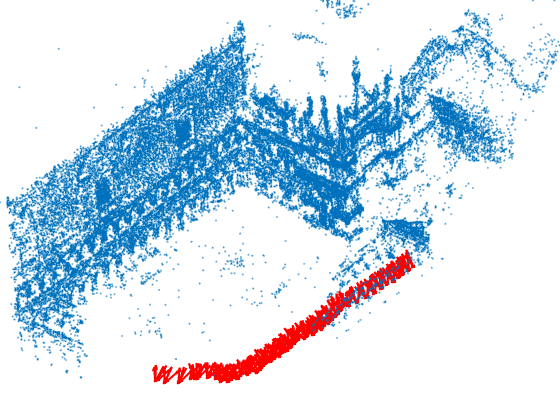}} & \eqref{eq:pOSE}& \textbf{1.074} & 0.073 & 0.483 &\\
&& \eqref{eq:rotpOSE} & 6.025 & \textbf{0.005} & \textbf{0.0438}\\ 
& &\\
\hline

\multirow{3}{*}{\begin{minipage}{\w}
   Nikolai I \\
   \textit{Cameras: \hspace{0.15cm} 98  \\
   3D points: \hspace{0.01cm} 37 739}
\end{minipage}} & \multirow{3}{*}{\includegraphics[height=14mm]{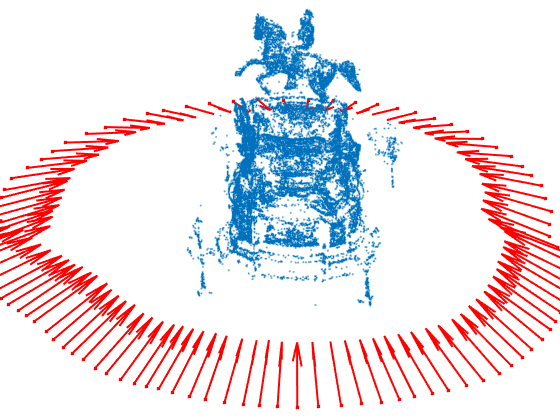}} & \eqref{eq:pOSE}& \textbf{1.074} & 0.073 & 0.483 &\\
&& \eqref{eq:rotpOSE} & 6.025 & \textbf{0.005} & \textbf{0.0438}\\ 
& &\\
\hline

\multirow{3}{*}{\begin{minipage}{\w}
    Council Chamber \\
    \textit{Cameras: \hspace{0.15cm} 176  \\
    3D points: \hspace{0.01cm} 110 819}
\end{minipage}} & \multirow{3}{*}{\includegraphics[height=14mm]{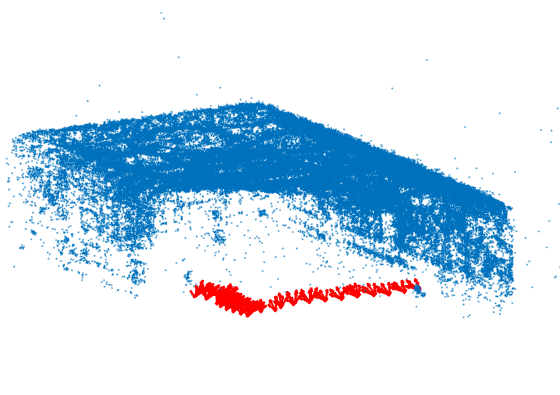}} & \eqref{eq:pOSE}& \textbf{3.235} & 0.057  & 0.491 &\\
&& \eqref{eq:rotpOSE} & 8.268 & \textbf{0.006} & \textbf{0.029}\\ 
& &\\
\hline

\multirow{3}{*}{\begin{minipage}{\w}
   LU Sphinx \\
   \textit{Cameras: \hspace{0.15cm} 70  \\
   3D points: \hspace{0.01cm} 13 566}
\end{minipage}} & \multirow{3}{*}{\includegraphics[height=14mm]{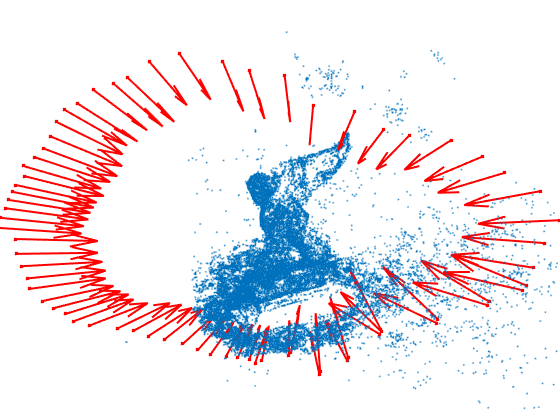}} & \eqref{eq:pOSE}& \textbf{0.163} & 0.033 & 0.151 &\\
&& \eqref{eq:rotpOSE} & 0.787 & \textbf{0.008} & \textbf{0.059}\\ 
& &\\
\hline

\multirow{3}{*}{\begin{minipage}{\w}
   Kamakura Hachimangu \\
   \textit{Cameras: \hspace{0.15cm} 99  \\
   3D points: \hspace{0.01cm} 24 980}
\end{minipage}} & \multirow{3}{*}{\includegraphics[height=14mm]{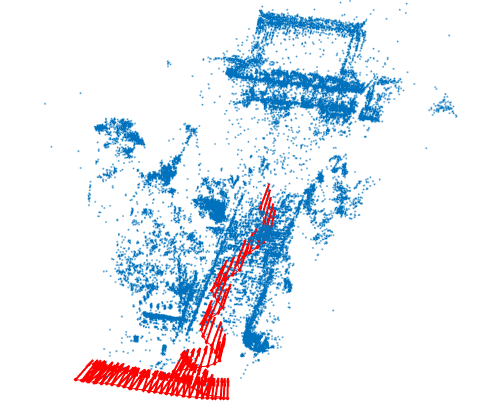}} & \eqref{eq:pOSE}& \textbf{0.131} & 0.005 & 0.027 &\\
&& \eqref{eq:rotpOSE} & 3.283 & \textbf{0.003} & \textbf{0.009}\\ 
& &\\
\hline

\multirow{3}{*}{\begin{minipage}{\w}
    Ahlstromer \\
    \textit{Cameras: \hspace{0.15cm} 41  \\
    3D points: \hspace{0.01cm} 2021}
\end{minipage}} & \multirow{3}{*}{\includegraphics[height=14mm]{recahlstromer.png}} & \eqref{eq:pOSE} & \textbf{0.003} & 0.013 & 0.034\\
&& \eqref{eq:rotpOSE} &    0.0493& \textbf{0.001} &  \textbf{0.002}\\ 
%&& $l_{diag}+l_{pOSE}$ & 0.731  &  0.004 & 0.011 \\ 
%&& $l_{rot}+l_{pOSE}$ (param) & 0.0496  & \textbf{0} & \textbf{0}\\ 
& &\\
\hline

\multirow{3}{*}{\begin{minipage}{\w}
    Park Gate \\
    \textit{Cameras: \hspace{0.15cm} 34  \\
    3D points: \hspace{0.01cm} 6 634}
\end{minipage}} & \multirow{3}{*}{\includegraphics[height=14mm]{rec_park_gate.png}} & \eqref{eq:pOSE} & \textbf{0.0223} & 0.006  & 0.0.023 \\
&& \eqref{eq:rotpOSE} & 0.231 & \textbf{0.002} & \textbf{0.009}\\ 
%&& $l_{diag}+l_{pOSE}$ & 1.590 & 0.022 & 0.078\\ 
%&& $l_{rot}+l_{pOSE}$  (param) & 0.24 & \textbf{0}& \textbf{0}\\ 
& &\\
\hline

\multirow{3}{*}{\begin{minipage}{\w}
    De Guerre \\
    \textit{Cameras: \hspace{0.15cm} 35  \\
    3D points: \hspace{0.01cm} 13 404}
\end{minipage}} & \multirow{3}{*}{\includegraphics[height=14mm]{rec_DeGuerre.png}} & \eqref{eq:pOSE} & \textbf{0.072} & 0.01 & 0.04 &\\
&& \eqref{eq:rotpOSE} & 0.362& \textbf{0.004} & \textbf{0.012}\\ 
%&& $l_{diag}+l_{pOSE}$ & 5.822  & 0.038 & 0.13 \\ 
%&& $l_{rot}+l_{pOSE}$ (param)& 10.062  & \textbf{0}& \textbf{0}\\ 
& &\\
\hline

\multirow{3}{*}{\begin{minipage}{\w}
    UWO \\
    \textit{Cameras: \hspace{0.15cm} 114  \\
    3D points: \hspace{0.01cm} 10 261}
\end{minipage}} & \multirow{3}{*}{\includegraphics[height=14mm]{uwo_rec}} & \eqref{eq:pOSE} & \textbf{0.055} & 0.15   & 0.625 &\\
&& \eqref{eq:rotpOSE} & 1.126 & \textbf{0.004} &  0.055\\ 
%&& $l_{diag}+l_{pOSE}$ & 7.34  & 0.01 & 0.05\\ 
%&& $l_{rot}+l_{pOSE}$ (param) & 1.166  & \textbf{0}& \textbf{0}\\ 
& & \\
\hline

\multirow{3}{*}{\begin{minipage}{\w}
    Togo \\
    \textit{Cameras: \hspace{0.15cm} 25  \\
    3D points: \hspace{0.01cm} 2 813}
\end{minipage}} & \multirow{3}{*}{\includegraphics[height=14mm]{rec_Togo.png}} & \eqref{eq:pOSE} & \textbf{0.007} & 0.008 & 0.02 &\\
&& \eqref{eq:rotpOSE} & 0.138 & \textbf{0.002} &\textbf{ 0.005}\\ 
%&& $l_{diag}+l_{pOSE}$ & 1.424  & 0.008 &0.027\\ 
%&& $l_{rot}+l_{pOSE}$ (param) &  1.551  & \textbf{0}& \textbf{0}\\ 
& & \\
\hline
\hline
		\end{tabular}
	\end{center}
	\caption{Method comparison with pOSE errors and upgradability measures for the optimal solutions. }
\end{table*}\label{Table_result}

\section{Conclusions}
In this paper we have presented an extension of the pOSE framework that provides near metric reconstructions. Our method combines pOSE with rotation averaging by incorporating relative rotation estimates into the objective function. 
Since the new error residuals are only invariant to similarity transformations the result is a visually accurate solution which does not need to be upgraded to conform to the real scene geometry. 

Our experimental evaluation shows that simple modifications of the pOSE framework, such as parameterization with rotations and penalizing deviation from the orthogonal constraints, lead to formulations that easily get stuck in local minima. In contrast our approach
leads to a well behaved formulation that converges to the right solution from random non-metric starting solutions with high probability.

{
    \small
    \bibliographystyle{ieeenat_fullname}
    \bibliography{main}
}

\end{document}

%% file: main.bib
@String(CVPR= {IEEE Conf. Comput. Vis. Pattern Recog.})

@String(ICCV= {Int. Conf. Comput. Vis.})

@String(ECCV= {Eur. Conf. Comput. Vis.})

@String(ICPR = {Int. Conf. Pattern Recog.})

@String(BMVC= {Brit. Mach. Vis. Conf.})

@String(ACCV  = {ACCV})

@String(CVPR  = {CVPR})

@String(ICCV  = {ICCV})

@String(ECCV  = {ECCV})

@String(ICPR  = {ICPR})

@String(BMVC  =	{BMVC})

@InProceedings{kasten_2019_ICCV,
author = {Kasten, Yoni and Geifman, Amnon and Galun, Meirav and Basri, Ronen},
title = {Algebraic Characterization of Essential Matrices and Their Averaging in Multiview Settings},
booktitle = {Proceedings of the IEEE/CVF International Conference on Computer Vision (ICCV)},
month = {October},
year = {2019}
}

@article{nasihatkon-etal-2015,
author = {Nasihatkon, Behrooz and Hartley, Richard and Trumpf, Jochen},
title = {A Generalized Projective Reconstruction Theorem and Depth Constraints for Projective Factorization},
year = {2015},
issue_date = {November  2015},
publisher = {Kluwer Academic Publishers},
address = {USA},
volume = {115},
number = {2},
issn = {0920-5691},
url = {https://doi.org/10.1007/s11263-015-0803-3},
doi = {10.1007/s11263-015-0803-3},
journal = {Int. J. Comput. Vision},
month = nov,
pages = {87–114},
numpages = {28}
}

@inproceedings{pollefeys1997stratified,
  title={A stratified approach to metric self-calibration},
  author={Pollefeys, Marc and Van Gool, Luc},
  booktitle={Proceedings of IEEE computer society conference on computer vision and pattern recognition},
  pages={407--412},
  year={1997},
  organization={IEEE}
}

@INPROCEEDINGS{hartley-1999,
  author={Hartley, R.I. and Hayman, E. and de Agapito, L. and Reid, I.},
  booktitle={Proceedings of the Seventh IEEE International Conference on Computer Vision}, 
  title={Camera calibration and the search for infinity}, 
  year={1999},
  volume={1},
  number={},
  pages={510-517 vol.1},
  doi={10.1109/ICCV.1999.791264}}

@article{faugeras1995stratification,
  title={Stratification of three-dimensional vision: projective, affine, and metric representations},
  author={Faugeras, Olivier},
  journal={Journal of the Optical Society of America A},
  volume={12},
  number={3},
  pages={465--484},
  year={1995},
  publisher={Optical Society of America}
}

@article{faugeras19983,
  title={3-d reconstruction of urban scenes from image sequences},
  author={Faugeras, Olivier and Robert, Luc and Laveau, St{\'e}phane and Csurka, Gabriella and Zeller, Cyril and Gauclin, Cyrille and Zoghlami, Imad},
  journal={Computer vision and image understanding},
  volume={69},
  number={3},
  pages={292--309},
  year={1998},
  publisher={Elsevier}
}

@article{fusiello-ivc-2000,
title = {Uncalibrated Euclidean reconstruction: a review},
journal = {Image and Vision Computing},
volume = {18},
number = {6},
pages = {555-563},
year = {2000},
issn = {0262-8856},
doi = {https://doi.org/10.1016/S0262-8856(99)00065-7},
url = {https://www.sciencedirect.com/science/article/pii/S0262885699000657},
author = {A. Fusiello}
}

@inproceedings{weber-etal-eccv-2024,
author = {Weber, Simon and Hong, Je Hyeong and Cremers, Daniel},
title = {Power Variable Projection for Initialization-Free Large-Scale Bundle Adjustment},
year = {2024},
isbn = {978-3-031-72623-1},
publisher = {Springer-Verlag},
address = {Berlin, Heidelberg},
url = {https://doi.org/10.1007/978-3-031-72624-8_7},
doi = {10.1007/978-3-031-72624-8_7},
booktitle = {Computer Vision – ECCV 2024: 18th European Conference, Milan, Italy, September 29–October 4, 2024, Proceedings, Part XIII},
pages = {111–126},
numpages = {16},
location = {Milan, Italy}
}

@InProceedings{Iglesias_2023_CVPR,
    author    = {Iglesias, Jos\'e Pedro and Nilsson, Amanda and Olsson, Carl},
    title     = {expOSE: Accurate Initialization-Free Projective Factorization Using Exponential Regularization},
    booktitle = {Proceedings of the IEEE/CVF Conference on Computer Vision and Pattern Recognition (CVPR)},
    month     = {June},
    year      = {2023},
    pages     = {8959-8968}
}

@INPROCEEDINGS{iglesias-olsson-iccv-2021,
  author={Iglesias, José Pedro and Olsson, Carl},
  booktitle={2021 IEEE/CVF International Conference on Computer Vision (ICCV)}, 
  title={Radial Distortion Invariant Factorization for Structure from Motion}, 
  year={2021},
  volume={},
  number={},
  pages={5886-5895},
  doi={10.1109/ICCV48922.2021.00585}}

@inproceedings{hong-fitzgibbon-iccv-2015,
author = {Hong, Je Hyeong and Fitzgibbon, Andrew},
title = {Secrets of Matrix Factorization: Approximations, Numerics, Manifold Optimization and Random Restarts},
booktitle = {Int. Conf. on Computer Vision},
year = {2015},
}

@inproceedings{hong-etal-eccv-2016,
  title={Projective Bundle Adjustment from Arbitrary Initialization Using the Variable Projection Method},
  author={Je Hyeong Hong and Christopher Zach and Andrew W. Fitzgibbon and Roberto Cipolla},
  booktitle={European Conf. on Computer Vision},
  year={2016}
}

@InProceedings{hong_2018,
author = {Hong, Je Hyeong and Zach, Christopher},
title = {pOSE: Pseudo Object Space Error for Initialization-Free Bundle Adjustment},
booktitle = {Proceedings of the IEEE Conference on Computer Vision and Pattern Recognition (CVPR)},
month = {June},
year = {2018}
}

@INPROCEEDINGS{hong_2017,  
author={J. H. {Hong} and C. {Zach} and A. {Fitzgibbon}},  booktitle={2017 IEEE Conference on Computer Vision and Pattern Recognition (CVPR)},   title={Revisiting the Variable Projection Method for Separable Nonlinear Least Squares Problems},   year={2017},  volume={},  number={},  pages={5939-5947},}

@inproceedings{rother-carlsson-eccv-2002,
author = {Rother, Carsten and Carlsson, Stefan},
title = {Linear Multi View Reconstruction with Missing Data},
year = {2002},
isbn = {3540437444},
publisher = {Springer-Verlag},
address = {Berlin, Heidelberg},
booktitle = {Proceedings of the 7th European Conference on Computer Vision-Part II},
pages = {309–324},
numpages = {16},
series = {ECCV '02}
}

@inproceedings{rother-carlsson-2001,
author = {Rother, Carsten and Carlsson, Stefan},
year = {2001},
month = {02},
pages = {42-50 vol.1},
title = {Linear Multi View Reconstruction and Camera Recovery},
volume = {1},
isbn = {0-7695-1143-0},
journal = {Proceedings of the IEEE International Conference on Computer Vision},
doi = {10.1109/ICCV.2001.937497}
}

@ARTICLE{kahl-hartley-tpami-2008,
  author={Kahl, Fredrik and Hartley, Richard},
  journal={IEEE Transactions on Pattern Analysis and Machine Intelligence}, 
  title={Multiple-View Geometry Under the {$L_\infty$}-Norm}, 
  year={2008},
  volume={30},
  number={9},
  pages={1603-1617},
  doi={10.1109/TPAMI.2007.70824}}

@INPROCEEDINGS{zhang-etal-2018,
  author={Zhang, Qianggong and Chin, Tat-Jun and Le, Huu Minh},
  booktitle={2018 IEEE/CVF Conference on Computer Vision and Pattern Recognition}, 
  title={A Fast Resection-Intersection Method for the Known Rotation Problem}, 
  year={2018},
  volume={},
  number={},
  pages={3012-3021},
  doi={10.1109/CVPR.2018.00318}}

@INPROCEEDINGS{cui-tan-iccv-2015,
  author={Cui, Zhaopeng and Tan, Ping},
  booktitle={2015 IEEE International Conference on Computer Vision (ICCV)}, 
  title={Global Structure-from-Motion by Similarity Averaging}, 
  year={2015},
  volume={},
  number={},
  pages={864-872},
  doi={10.1109/ICCV.2015.105}}

@INPROCEEDINGS{kennedy-etal-2012,
  author={Kennedy, Ryan and Daniilidis, Kostas and Naroditsky, Oleg and Taylor, Camillo J.},
  booktitle={2012 IEEE/RSJ International Conference on Intelligent Robots and Systems}, 
  title={Identifying maximal rigid components in bearing-based localization}, 
  year={2012},
  volume={},
  number={},
  pages={194-201},
  doi={10.1109/IROS.2012.6386132}}

@inproceedings{cui-etal-bmvc-2015,
  author = {Cui, Zhaopeng and Jiang, Nianjuan and Tang, Chengzhou and Tan, Ping},
  booktitle = {BMVC},
  editor = {Xie, Xianghua and Jones, Mark W. and Tam, Gary K. L.},
  ee = {http://dx.doi.org/10.5244/C.29.46},
  isbn = {1-901725-53-7},
  pages = {46.1-46.13},
  publisher = {BMVA Press},
  title = {Linear Global Translation Estimation with Feature Tracks.},
  year = 2015,
}

@INPROCEEDINGS{hainan-etal-2016,
  author={Cui, Hainan and Shen, Shuhan and Hu, Zhanyi},
  booktitle={2016 23rd International Conference on Pattern Recognition (ICPR)}, 
  title={Robust global translation averaging with feature tracks}, 
  year={2016},
  volume={},
  number={},
  pages={3727-3732},
  doi={10.1109/ICPR.2016.7900214}}

@inproceedings{latit-govindu-2022,
author = {Manam, Lalit and Govindu, Venu Madhav},
title = {Correspondence Reweighted Translation Averaging},
year = {2022},
isbn = {978-3-031-19826-7},
publisher = {Springer-Verlag},
address = {Berlin, Heidelberg},
url = {https://doi.org/10.1007/978-3-031-19827-4_4},
doi = {10.1007/978-3-031-19827-4_4},
booktitle = {Computer Vision – ECCV 2022: 17th European Conference, Tel Aviv, Israel, October 23–27, 2022, Proceedings, Part XXXIII},
pages = {56–72},
numpages = {17},
location = {Tel Aviv, Israel}
}

@INPROCEEDINGS{martinec-pajdla-2007,
  author={Martinec, Daniel and Pajdla, Tomas},
  booktitle={2007 IEEE Conference on Computer Vision and Pattern Recognition}, 
  title={Robust Rotation and Translation Estimation in Multiview Reconstruction}, 
  year={2007},
  volume={},
  number={},
  pages={1-8},
  doi={10.1109/CVPR.2007.383115}}

@INPROCEEDINGS{arie-nachimson-etal-2012,
  author={Arie-Nachimson, Mica and Kovalsky, Shahar Z. and Kemelmacher-Shlizerman, Ira and Singer, Amit and Basri, Ronen},
  booktitle={2012 Second International Conference on 3D Imaging, Modeling, Processing, Visualization \& Transmission}, 
  title={Global Motion Estimation from Point Matches}, 
  year={2012},
  volume={},
  number={},
  pages={81-88},
  doi={10.1109/3DIMPVT.2012.46}}

@article{pollefeys-etal-ijcv-2008,
author = {Pollefeys, M. and Nist\'{e}r, D. and Frahm, J. -M. and Akbarzadeh, A. and Mordohai, P. and Clipp, B. and Engels, C. and Gallup, D. and Kim, S. -J. and Merrell, P. and Salmi, C. and Sinha, S. and Talton, B. and Wang, L. and Yang, Q. and Stew\'{e}nius, H. and Yang, R. and Welch, G. and Towles, H.},
title = {Detailed Real-Time Urban 3D Reconstruction from Video},
year = {2008},
issue_date = {July      2008},
publisher = {Kluwer Academic Publishers},
address = {USA},
volume = {78},
number = {2–3},
issn = {0920-5691},
url = {https://doi.org/10.1007/s11263-007-0086-4},
doi = {10.1007/s11263-007-0086-4},
journal = {Int. J. Comput. Vision},
month = jul,
pages = {143–167},
numpages = {25}
}

@inproceedings{Snavely_SIGGRAPH06,
  address = {New York, NY, USA},
  author = {Snavely, Noah and Seitz, Steven M. and Szeliski, Richard},
  booktitle = {SIGGRAPH Conference Proceedings},
  isbn = {1-59593-364-6},
  pages = {835--846},
  publisher = {ACM Press},
  title = {Photo tourism: Exploring photo collections in 3D},
  url = {http://phototour.cs.washington.edu/},
  year = 2006
}

@inproceedings{engels-etal-2006,
author = {Engels, Chris and Stewénius, Henrik and Nistér, David},
year = {2006},
month = {01},
pages = {},
title = {Bundle adjustment rules}
}

@INPROCEEDINGS{kai-etel-2007,
  author={Ni, Kai and Steedly, Drew and Dellaert, Frank},
  booktitle={2007 IEEE 11th International Conference on Computer Vision}, 
  title={Out-of-Core Bundle Adjustment for Large-Scale 3D Reconstruction}, 
  year={2007},
  volume={},
  number={},
  pages={1-8},
  doi={10.1109/ICCV.2007.4409085}}

@inproceedings{agarwal-etal-eccv-2010,
author = {Agarwal, Sameer and Snavely, Noah and Seitz, Steven M. and Szeliski, Richard},
title = {Bundle adjustment in the large},
year = {2010},
isbn = {3642155510},
publisher = {Springer-Verlag},
address = {Berlin, Heidelberg},
booktitle = {Proceedings of the 11th European Conference on Computer Vision: Part II},
pages = {29–42},
numpages = {14},
location = {Heraklion, Crete, Greece},
series = {ECCV'10}
}

@article{agarwal-etal-2011,
author = {Agarwal, Sameer and Furukawa, Yasutaka and Snavely, Noah and Simon, Ian and Curless, Brian and Seitz, Steven M. and Szeliski, Richard},
title = {Building Rome in a day},
year = {2011},
issue_date = {October 2011},
publisher = {Association for Computing Machinery},
address = {New York, NY, USA},
volume = {54},
number = {10},
issn = {0001-0782},
url = {https://doi.org/10.1145/2001269.2001293},
doi = {10.1145/2001269.2001293},
journal = {Commun. ACM},
month = oct,
pages = {105–112},
numpages = {8}
}

@inproceedings{byrod-eccv-2010,
author = {Byr\"{o}d, Martin and \r{A}str\"{o}m, Kalle},
title = {Conjugate gradient bundle adjustment},
year = {2010},
isbn = {3642155510},
publisher = {Springer-Verlag},
address = {Berlin, Heidelberg},
booktitle = {Proceedings of the 11th European Conference on Computer Vision: Part II},
pages = {114–127},
numpages = {14},
location = {Heraklion, Crete, Greece},
series = {ECCV'10}
}

@ARTICLE{cornelis2004,
  author={Cornelis, K. and Verbiest, F. and Van Gool, L.},
  journal={IEEE Transactions on Pattern Analysis and Machine Intelligence}, 
  title={Drift detection and removal for sequential structure from motion algorithms}, 
  year={2004},
  volume={26},
  number={10},
  pages={1249-1259},
  doi={10.1109/TPAMI.2004.85}}

@book{hartley-zisserman-2003,
  address = {New York, NY, USA},
  author = {Hartley, Richard and Zisserman, Andrew},
  description = {Multiple View Geometry in Computer Vision},
  edition = 2,
  isbn = {0521540518},
  publisher = {Cambridge University Press},
  title = {Multiple View Geometry in Computer Vision},
  year = 2003
}

@inproceedings{moulon2013,
  TITLE = {{Global Fusion of Relative Motions for Robust, Accurate and Scalable Structure from Motion}},
  AUTHOR = {Moulon, Pierre and Monasse, Pascal and Marlet, Renaud},
  URL = {https://enpc.hal.science/hal-00873504},
  BOOKTITLE = {{Proceedings of IEEE International Conference on Computer Vision}},
  ADDRESS = {Sydney, Australia},
  PAGES = {to appear},
  YEAR = {2013},
  MONTH = Dec,
  HAL_ID = {hal-00873504},
  HAL_VERSION = {v1},
}

@misc{pan2024,
      title={Global Structure-from-Motion Revisited}, 
      author={Linfei Pan and Dániel Baráth and Marc Pollefeys and Johannes L. Schönberger},
      year={2024},
      eprint={2407.20219},
      archivePrefix={arXiv},
      primaryClass={cs.CV},
      url={https://arxiv.org/abs/2407.20219}, 
}

@INPROCEEDINGS{olsson-etal-cvpr-2010,
  author={Olsson, Carl and Eriksson, Anders and Hartley, Richard},
  booktitle={2010 IEEE Computer Society Conference on Computer Vision and Pattern Recognition}, 
  title={Outlier removal using duality}, 
  year={2010},
  volume={},
  number={},
  pages={1450-1457},
  doi={10.1109/CVPR.2010.5539800}}

@inproceedings{triggs1999,
	author = {Triggs, Bill and McLauchlan, Philip F. and Hartley, Richard I. and Fitzgibbon, Andrew W.},
	title = {Bundle Adjustment - A Modern Synthesis},
	year = {1999},
	isbn = {3540679731},
	publisher = {Springer-Verlag},
	address = {Berlin, Heidelberg},
	booktitle = {Proceedings of the International Workshop on Vision Algorithms: Theory and Practice},
	pages = {298–372},
	numpages = {75},
	series = {ICCV '99}
}

@InProceedings{fredriksson2012,
author="Fredriksson, Johan
and Olsson, Carl",
editor="Lee, Kyoung Mu
and Matsushita, Yasuyuki
and Rehg, James M.
and Hu, Zhanyi",
title="Simultaneous Multiple Rotation Averaging Using Lagrangian Duality",
booktitle="Computer Vision -- ACCV 2012",
year="2012",
publisher="Springer Berlin Heidelberg",
address="Berlin, Heidelberg",
pages="245--258",
isbn="978-3-642-37431-9"
}

@article{nister2004,
	title={An efficient solution to the five-point relative pose problem},
	author={David Nist{\'e}r},
	journal={IEEE Transactions on Pattern Analysis and Machine Intelligence},
	year={2004},
	volume={26},
	pages={756-770},
	url={https://api.semanticscholar.org/CorpusID:886598}
}

@InProceedings{olsson2011,
	author="Olsson, Carl
	and Enqvist, Olof",
	editor="Heyden, Anders
	and Kahl, Fredrik",
	title="Stable Structure from Motion for Unordered Image Collections",
	booktitle="Image Analysis",
	year="2011",
	publisher="Springer Berlin Heidelberg",
	address="Berlin, Heidelberg",
	pages="524--535",
	isbn="978-3-642-21227-7"
}

@INPROCEEDINGS{enqvist2011,
	author={Enqvist, Olof and Kahl, Fredrik and Olsson, Carl},
	booktitle={2011 IEEE International Conference on Computer Vision Workshops (ICCV Workshops)}, 
	title={Non-sequential structure from motion}, 
	year={2011},
	volume={},
	number={},
	pages={264-271},
	doi={10.1109/ICCVW.2011.6130252}}

@article{briales2017,
	author = {Briales, Jesus and González-Jiménez, Javier},
	year = {2017},
	month = {06},
	pages = {1-1},
	title = {Cartan-Sync: Fast and Global SE(d)-Synchronization},
	volume = {PP},
	journal = {IEEE Robotics and Automation Letters},
	doi = {10.1109/LRA.2017.2718661}
}

@article{wang2013,
	author = {Wang, Lanhui and Singer, Amit},
	title = "{Exact and stable recovery of rotations for robust synchronization}",
	journal = {Information and Inference: A Journal of the IMA},
	volume = {2},
	number = {2},
	pages = {145-193},
	year = {2013},
	month = {12},
	issn = {2049-8764},
	doi = {10.1093/imaiai/iat005},
	url = {https://doi.org/10.1093/imaiai/iat005},
	eprint = {https://academic.oup.com/imaiai/article-pdf/2/2/145/1958172/iat005.pdf},
}

@article{singer2011,
	title = {Angular synchronization by eigenvectors and semidefinite programming},
	journal = {Applied and Computational Harmonic Analysis},
	volume = {30},
	number = {1},
	pages = {20-36},
	year = {2011},
	issn = {1063-5203},
	doi = {https://doi.org/10.1016/j.acha.2010.02.001},
	url = {https://www.sciencedirect.com/science/article/pii/S1063520310000205},
	author = {A. Singer},
}

@article{Carlone2015b,
	author = {Carlone, Luca and Tron, Roberto and Daniilidis, Kostas and Dellaert, Frank},
	year = {2015},
	month = {06},
	pages = {4597-4604},
	title = {Initialization techniques for 3D SLAM: A survey on rotation estimation and its use in pose graph optimization},
	volume = {2015},
	journal = {Proceedings - IEEE International Conference on Robotics and Automation},
	doi = {10.1109/ICRA.2015.7139836}
}

@InProceedings{Moreira2021,
	author    = {Moreira, Gabriel and Marques, Manuel and Costeira, Jo\~ao Paulo},
	title     = {Rotation Averaging in a Split Second: A Primal-Dual Method and a Closed-Form for Cycle Graphs},
	booktitle = {Proceedings of the IEEE/CVF International Conference on Computer Vision (ICCV)},
	month     = {October},
	year      = {2021},
	pages     = {5452-5460}
}

@InProceedings{parra2021,
	author    = {Parra, Alvaro and Chng, Shin-Fang and Chin, Tat-Jun and Eriksson, Anders and Reid, Ian},
	title     = {Rotation Coordinate Descent for Fast Globally Optimal Rotation Averaging},
	booktitle = {Proceedings of the IEEE/CVF Conference on Computer Vision and Pattern Recognition (CVPR)},
	month     = {June},
	year      = {2021},
	pages     = {4298-4307}
}

@article{rosen2021,
	author = {Rosen, David and Doherty, Kevin and Espinoza, Antonio and Leonard, John},
	year = {2021},
	month = {05},
	pages = {},
	title = {Advances in Inference and Representation for Simultaneous Localization and Mapping},
	volume = {4},
	journal = {Annual Review of Control, Robotics, and Autonomous Systems},
	doi = {10.1146/annurev-control-072720-082553}
}

@InProceedings{dai2009,
	author="Dai, Yuchao
	and Trumpf, Jochen
	and Li, Hongdong
	and Barnes, Nick
	and Hartley, Richard",
	title="Rotation Averaging with Application to Camera-Rig Calibration",
	booktitle="Computer Vision -- ACCV 2009",
	year="2010",
}

@InProceedings{chen2021,
	author    = {Chen, Yu and Zhao, Ji and Kneip, Laurent},
	title     = {Hybrid Rotation Averaging: A Fast and Robust Rotation Averaging Approach},
	booktitle = {Proceedings of the IEEE/CVF Conference on Computer Vision and Pattern Recognition (CVPR)},
	month     = {June},
	year      = {2021},
	pages     = {10358-10367}
}

@article{hartley2011,
	title={L1 rotation averaging using the Weiszfeld algorithm},
	author={Richard I. Hartley and Khurrum Aftab and Jochen Trumpf},
	journal={CVPR 2011},
	year={2011},
	pages={3041-3048},
}

@article{hartley2013,
	Author = {Hartley, Richard and Trumpf, Jochen and Dai, Yuchao and Li, Hongdong},
	ISSN = {09205691},
	Journal = {International Journal of Computer Vision},
	Number = {3},
	Pages = {267 - 305},
	Title = {Rotation Averaging.},
	Volume = {103},
	Year = {2013},
}

@INPROCEEDINGS{chitturi2021,
	author={Sidhartha, Chitturi and Govindu, Venu Madhav},
	booktitle={2021 International Conference on 3D Vision (3DV)}, 
	title={It Is All In The Weights: Robust Rotation Averaging Revisited}, 
	year={2021},
	volume={},
	number={},
	pages={1134-1143},
	doi={10.1109/3DV53792.2021.00121}}

@ARTICLE{chatterjee2017,
	author                  = {A. Chatterjee and V. Govindu},
	journal                 = {IEEE Transactions on Pattern Analysis and Machine Intelligence},
	title                   = {Robust Relative Rotation Averaging},
	year                    = {2017},
	volume                  = {40},
	number                  = {4},
	pages                   = {958--972},
	doi                     = {10.1109/TPAMI.2017.2693984},
	url                     = {https://ee.iisc.ac.in/cvlab/papers/robustrelrotavg.pdf},
	Comment                 = {<a href="https://ee.iisc.ac.in/cvlab/research/rotaveraging/">[project page]</a>},
	ISSN                    = {0162-8828}  
}

@inproceedings{wilson2016,
	author = {Wilson, Kyle and Bindel, David and Snavely, Noah},
	booktitle = {Proceedings of ECCV 2016},
	title = {When is Rotations Averaging Hard?},
	month = oct,
	year = {2016}
}

@ARTICLE{eriksson2021,
	author={Eriksson, Anders and Olsson, Carl and Kahl, Fredrik and Chin, Tat-Jun},
	journal={IEEE Transactions on Pattern Analysis and Machine Intelligence}, 
	title={Rotation Averaging with the Chordal Distance: Global Minimizers and Strong Duality}, 
	year={2021},
	volume={43},
	number={1},
	pages={256-268},
	doi={10.1109/TPAMI.2019.2930051}}

@inproceedings{Zhang2023,
	title = "Revisiting Rotation Averaging: Uncertainties and Robust Losses",
	author = "Ganlin Zhang and Viktor Larsson and Daniel Barath",
	year = "2023",
	doi = "10.1109/CVPR52729.2023.01651",
	series = "Proceedings of the IEEE Computer Society Conference on Computer Vision and Pattern Recognition",
	publisher = "IEEE Computer Society",
	pages = "17215--17224",
	booktitle = "Proceedings - 2023 IEEE/CVF Conference on Computer Vision and Pattern Recognition, CVPR 2023",
	address = "United States",
	note = "2023 IEEE/CVF Conference on Computer Vision and Pattern Recognition, CVPR 2023 ; Conference date: 18-06-2023 Through 22-06-2023",
}

@article{dellaert2020,
	author       = {Frank Dellaert and
	David M. Rosen and
	Jing Wu and
	Robert E. Mahony and
	Luca Carlone},
	title        = {Shonan Rotation Averaging: Global Optimality by Surfing SO(p)\({}^{\mbox{n}}\)},
	journal      = {CoRR},
	volume       = {abs/2008.02737},
	year         = {2020},
	url          = {https://arxiv.org/abs/2008.02737},
	eprinttype    = {arXiv},
	eprint       = {2008.02737},
}

@INPROCEEDINGS{wilson2020,
	author={Wilson, Kyle and Bindel, David},
	booktitle={2020 IEEE/CVF Conference on Computer Vision and Pattern Recognition (CVPR)}, 
	title={On the Distribution of Minima in Intrinsic-Metric Rotation Averaging}, 
	year={2020},
	volume={},
	number={},
	pages={6030-6038},
	doi={10.1109/CVPR42600.2020.00607}}

@inproceedings{schoenberger2016sfm,
    author={Sch\"{o}nberger, Johannes Lutz and Frahm, Jan-Michael},
    title={Structure-from-Motion Revisited},
    booktitle={Conference on Computer Vision and Pattern Recognition (CVPR)},
    year={2016},
}

@article{hartley-sturm-1997,
  author = {Hartley, Richard I. and Sturm, Peter},
  description = {Triangulation - ScienceDirect},
  doi = {https://doi.org/10.1006/cviu.1997.0547},
  issn = {1077-3142},
  journal = {Computer Vision and Image Understanding},
  number = 2,
  pages = {146 - 157},
  title = {Triangulation},
  url = {http://www.sciencedirect.com/science/article/pii/S1077314297905476},
  volume = 68,
  year = 1997
}
